\pdfoutput=1
\documentclass[11pt]{article}

\usepackage[preprint]{acl}

\usepackage{times}
\usepackage{latexsym}

\usepackage[T1]{fontenc}
\usepackage[utf8]{inputenc}

\usepackage{microtype}

\usepackage{inconsolata}

\usepackage{graphicx}
\usepackage{subcaption}
\usepackage{booktabs}
\usepackage{tikz}
\usepackage{tcolorbox}
\usetikzlibrary{arrows.meta, positioning}
\usepackage{caption}
\usepackage{amssymb}
\usepackage{amsmath}
\usepackage{array}
\usepackage{lipsum} % For placeholder text
\usepackage{enumitem}

\title{Rating Roulette: Self-Inconsistency in LLM-As-A-Judge Frameworks}

\author{
Rajarshi Haldar \and Julia Hockenmaier \\
University of Illinois Urbana-Champaign \\
\texttt{\{rhaldar2, juliahmr\}@illinois.edu}
}

\tcbset{
  boxsep=1pt,
  top=1pt,
  bottom=1pt,
  left=4pt,
  right=4pt,
}

\begin{document}
\maketitle
\begin{abstract}
As Natural Language Generation (NLG) continues to be widely adopted, properly assessing it has become quite difficult. Lately, using large language models (LLMs) for evaluating these generations has gained traction, as they tend to align more closely with human preferences than conventional n-gram or embedding-based metrics. In our experiments, we show that LLM judges have low intra-rater reliability in their assigned scores across different runs. This variance makes their ratings inconsistent, almost arbitrary in the worst case, making it difficult to measure how good their judgments actually are. We quantify this inconsistency across different NLG tasks and benchmarks and see if judicious use of LLM judges can still be useful following proper guidelines.
\end{abstract}

\section{Introduction}
As Natural Language Generation (NLG) becomes more prevalent in a wide variety of applications like automated journalism, customer service chatbots, language translation and content summarization, proper evaluation and measurement of alignment with human preference are critical to improve user trust and system utility. Although traditionally used automatic metrics like BLEU~\citep{papineni-etal-2002-bleu} and BERTScore~\citep{bert-score} work well with tasks like translation when multiple references are available, they fail to work in more open-ended tasks like summarization or general chatbot settings.

To automatically evaluate LLMs in these settings, LLM-as-a-judge~\citep{li2024llmsasjudgescomprehensivesurveyllmbased, gu2025surveyllmasajudge} has emerged as an automatic, scalable alternative to manual evaluation. The common practice is to prompt an LLM to evaluate natural language generations. The ratings obtained are then verified by comparing with ratings assigned by human judges, which are taken as the gold standard. This comparison is commonly done through computing exact match for categorical labels or correlation for numeric or ordinal scales~\citep{liu-etal-2023-g,thakur2025judgingjudgesevaluatingalignment}.

However, what is often missing from these studies is any measure of self-reliability or intra-rater reliability of both LLM and human judges. We define self-reliability as the agreement of a judge with itself over multiple runs with the same settings (prompt and hyperparameters in case of LLMs).

Meanwhile, self-reliability data is completely missing from human annotations in existing benchmarks. In some cases, annotations from multiple human judges exist, but they only give us information about the inter-rater reliability instead, which is also frequently lower than the conventionally agreed upon agreement thresholds~\citep{falke-etal-2019-ranking,summeval,pagnoni-etal-2021-understanding}. Meanwhile, agreement between LLM and human judges is usually computed using metrics like correlation and accuracy, instead of using metrics specifically designed for measuring agreement like Krippendorff's Alpha. This can often lead to an overestimation of agreement since those metrics do not account for chance agreement~\citep{thakur-etal-2025-judging}.

% \subsection{Tasks and Benchmarks}
NLG can be evaluated using different metrics on a variety of tasks such as \textbf{machine translation}, \textbf{dialog generation} and \textbf{summarization}. In our experiments, we first study the simplest case of evaluating summarization --- assigning a \textbf{binary label} to a summary in the \textbf{SummaC} benchmark~\citep{laban-etal-2022-summac}. Next, we look at evaluating summaries in more complex scenarios, using a \textbf{Likert rating scale} of 1 to 5 to rate a summary on several metrics like coherence, consistency, fluency and relevance. We can use the \textbf{SummEval} benchmark for this purpose. Finally, we look at evaluating a different task of NLG, where we use a judge LLM to \textbf{rank} multi-turn conversations from two competing LLMs in the \textbf{MT-bench} benchmark~\citep{zheng2023judgingllmasajudgemtbenchchatbot}.

In this work, we make several key contributions to understanding the reliability of LLM-generated ratings in NLG evaluation. First, show that ratings output by LLMs have low agreement over multiple runs with the same prompt. Second, we show that turning off any sampling to make an LLM always output the same rating hurts performance measured by agreement with human judgment. Third, we find that this phenomenon persists across multiple tasks and benchmarks related to NLG, indicating that this is a widespread problem that needs to be addressed. Finally, we discuss some recommendations to conduct more robust NLG evaluations.

\section{Background}
\paragraph{Evaluating Natural Language Generation} The gold standard for NLG evaluation has relied on \textbf{human-centric evaluations}, where human judges assess generated text. However, this is time and cost-intensive, prone to unreliability and biases~\citep{humevalflaw}, and existing studies report low inter-annotator agreement or omit them altogether. For example, \citet{celikyilmaz2021evaluationtextgenerationsurvey} found that only 18\% of 135 surveyed papers included agreement analysis. Furthermore, leaderboards like Chatbot Arena~\citep{chiang2024chatbotarenaopenplatform} employing crowdsourced evaluations have inequities leading to an uneven playing field~\citep{singh2025leaderboardillusion}. Unlike human evaluations, automatic metrics like BLEU~\citep{papineni-etal-2002-bleu}, ROUGE~\citep{lin-2004-rouge}, and METEOR~\citep{banerjee-lavie-2005-meteor}, and model-based metrics like BERTScore~\citep{bert-score} and BLEURT~\citep{sellam-etal-2020-bleurt} are faster, less subjective and more scalable. However, these metrics often fail to fully capture human notions of quality, especially for subjective tasks.

\paragraph{LLM-as-a-judge}
Large language models can serve as judges by assigning scores, rankings, or labels to generated outputs, a paradigm surveyed in recent works~\citep{li2024llmsasjudgescomprehensivesurveyllmbased,gu2025surveyllmasajudge}. LLM judges have been applied across 20 NLP tasks~\citep{bavaresco2024llmsinsteadhumanjudges} and integrated into AI-assisted human evaluation~\citep{ashktorab2024aligninghumanllmjudgments}. Notably, GPT-4 with chain-of-thought prompts aligns more closely with human judgments than conventional automatic metrics on NLG tasks~\citep{liu-etal-2023-g,openai2024gpt4technicalreport,cot}. Evaluation typically involves comparing LLM outputs to human judgments via correlation~\citep{liu-etal-2023-g,xiao-etal-2023-evaluating-evaluation,bavaresco2024llmsinsteadhumanjudges} or percentage agreement~\citep{zheng2023judgingllmasajudgemtbenchchatbot} on benchmarks such as JudgeBench~\citep{tan2024judgebenchbenchmarkevaluatingllmbased}. Challenges include bias~\citep{zheng2023judgingllmasajudgemtbenchchatbot,li2025generationjudgmentopportunitieschallenges}, uncertainity in judgments~\citep{wagner2024blackboxuncertaintyquantificationmethod}, adversarial vulnerability~\citep{gu2025surveyllmasajudge}, and domain-specific performance gaps~\citep{tseng2024expertlevellanguagemodelsexpertlevel}, in addition to existing problems of LLM generations like prompt sensitivity~\citep{mizrahi-etal-2024-state}.  To enhance reliability, methods like panel-based evaluation (PoLL)~\citep{verga2024replacingjudgesjuriesevaluating} and comparative studies of fine-tuning versus GPT-4 prompting~\citep{huang2024empiricalstudyllmasajudgellm,openai2024gpt4technicalreport} have been proposed. There is another challenge, high variability in LLM outputs leading to inconsistencies in LLM judges.

\paragraph{Variability in LLM Judgments} Due to the stochastic nature of LLMs, they give different outputs when run on the same prompt multiple times. While this is by design, it should not alter the ratings assigned by an LLM judge to the same rating. ~\citet{chiang-lee-2023-large} observed some variability in ratings produced by LLM judges on the WritingPrompts~\citep{fan-etal-2018-hierarchical} dataset. ~\citet{liu-etal-2023-g} took advantage of this variability in GPT-4 by setting temperature to 1 and sampling 20 times to get multiple scores and then getting the final rating by normalizing all the scores by their probabilities. In our work, we go further to analyze the implications of high variability in assigned judgments, discuss metrics to measure this variability, study the extent of this issue across different tasks and benchmarks, and explore whether we would get better results by turning off the variability by setting the temperature to zero. We frame this variability as \textbf{self-reliability} or \textbf{intra-rater reliability}.

\paragraph{Self-reliability}
Also called intra-rater reliability, this measures the consistency of a single evaluator's judgments across repeated assessments, as opposed to inter-rater reliability. Common metrics include the intraclass correlation coefficient (ICC) for continuous scales~\citep{iccpaper,Bent2014}, Cohen's Kappa for categorical ratings, and Krippendorff's Alpha for ordinal, interval, or ratio data with missing values~\citep{Krippendorff2011ComputingKA,Wietholter2023}. Reporting intra-rater reliability is standard in domains like essay grading~\citep{irr_essay}, physical therapy~\citep{Mischiati2015}, and clinical diagnostics~\citep{Krolikowska2023}, and its joint consideration with inter-rater reliability is urged by ~\citet{harvey_irr}. Yet, NLP annotation and LLM-evaluation studies frequently omit self-reliability metrics~\citep{abercrombie-etal-2023-temporal,abercrombie2023consistencykeydisentanglinglabel}, a deficiency we confirm in our LLM-as-a-judge experiments, where models exhibit unstable self-judgments.

% \section{Experiments Evaluating Natural Language Generation with LLMs}

\section{Experiments}
We first consider a simple framework for NLG evaluation, binary classification of machine-generated summaries. A good candidate for this is \textbf{SummaC}, a factual consistency detection benchmark.

Following this, we make a deeper dive into the performance of LLM judges on the \textbf{SummEval} benchmark~\citep{summeval}. Compared to the binary labels in the SummaC benchmark, the summaries in this benchmark are evaluated by multiple human raters on a Likert scale across multiple metrics.

\citet{zheng2023judgingllmasajudgemtbenchchatbot} introduced the MT-bench benchmark, which evaluates an LLM's multi-turn conversational and instruction-following ability. While related to NLG, it is a very different task compared to summarization.

\subsection{Datasets}
%%%%% SummaC Example Starts Here
\begin{figure}[ht]
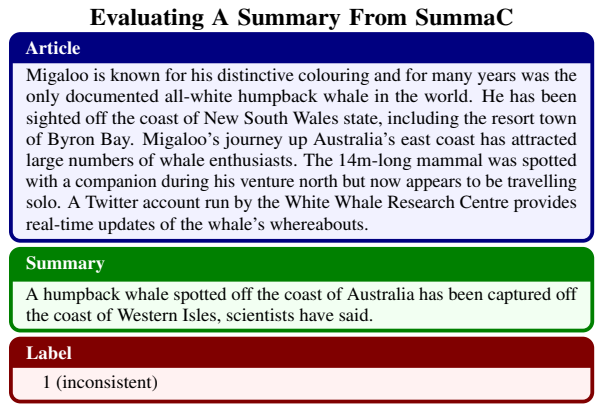

\centering
\begin{tcolorbox}[width=\linewidth, colback=white, colframe=white, halign=center, fontupper=\footnotesize\bfseries, 
boxsep=0pt, top=1pt, bottom=1pt]
% caption
Evaluating A Summary From SummaC
\end{tcolorbox}
\vspace{-15pt}
% article header
\begin{tcolorbox}[width=\linewidth, colback=blue!5, colframe=blue!50!black, title=Article, 
boxsep=1pt, top=1pt, bottom=1pt, toptitle=1pt, bottomtitle=1pt,
fonttitle=\scriptsize\bfseries]
\scriptsize
% article
Migaloo is known for his distinctive colouring and for many years was the only documented all-white humpback whale in the world. He has been sighted off the coast of New South Wales state, including the resort town of Byron Bay. Migaloo's journey up Australia's east coast has attracted large numbers of whale enthusiasts. The 14m-long mammal was spotted with a companion during his venture north but now appears to be travelling solo. A Twitter account run by the White Whale Research Centre provides real-time updates of the whale's whereabouts.
\end{tcolorbox}
\vspace{-12pt}
% summary header
\begin{tcolorbox}[colback=green!5, colframe=green!50!black, title=Summary, 
boxsep=1pt, top=1pt, bottom=1pt, toptitle=1pt, bottomtitle=1pt,
fonttitle=\scriptsize\bfseries]
\scriptsize
% summary
A humpback whale spotted off the coast of Australia has been captured off the coast of Western Isles, scientists have said.
\end{tcolorbox}
\vspace{-12pt}
% label header
\begin{tcolorbox}[width=\linewidth, colback=red!5, colframe=red!50!black, title=Label, 
boxsep=1pt, top=1pt, bottom=1pt, toptitle=1pt, bottomtitle=1pt,
fonttitle=\scriptsize\bfseries]
\scriptsize
\begin{tabular}{ll}
% label
1 (inconsistent)
\end{tabular}
\end{tcolorbox}

\caption{Annotating a summary of an article in SummaC with 1 (inconsistent).}
\label{fig:summac-evaluation-example}
\end{figure}
%%%%% SummaC Example Ends Here

\paragraph{SummaC}
The SummaC benchmark~\citep{laban-etal-2022-summac} evaluates factual consistency in text summarization by requiring judges to label summaries as either consistent or inconsistent with their source documents (example in Figure~\ref{fig:summac-evaluation-example}). It unifies six datasets, CoGenSumm~\citep{falke-etal-2019-ranking}, XSumFaith~\citep{maynez-etal-2020-faithfulness}, Polytope~\citep{huang-etal-2020-achieved}, FactCC~\citep{kryscinski-etal-2020-evaluating}, SummEval~\citep{summeval}, and FRANK~\citep{pagnoni-etal-2021-understanding}, into a binary classification task. Standard validation/test splits are used where available; otherwise, splits are created as needed. It is licensed under the Apache License Version 2.0. Dataset statistics are provided in Table~\ref{tab:summac-stats} in Appendix~\ref{app:summac_stats}.

%%%%% SummEval Example Starts Here
\begin{figure}[ht]
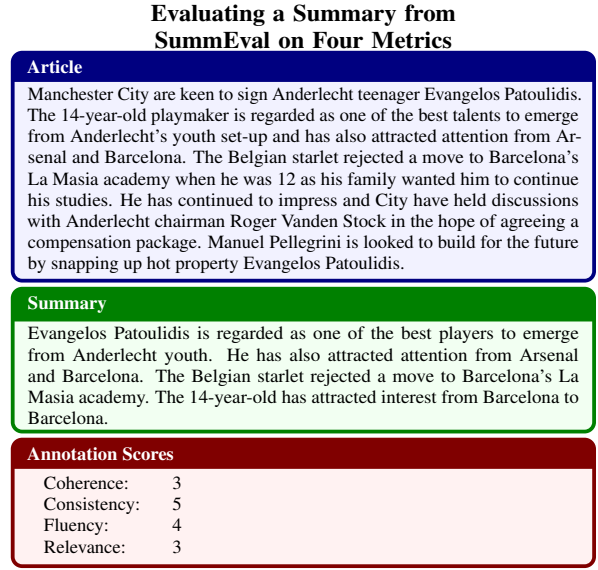

\centering
\begin{tcolorbox}[width=\linewidth, colback=white, colframe=white, halign=center, fontupper=\footnotesize\bfseries, 
boxsep=0pt, top=1pt, bottom=1pt]
Evaluating a Summary from SummEval on Four Metrics
\end{tcolorbox}
\vspace{-15pt}
\begin{tcolorbox}[width=\linewidth, colback=blue!5, colframe=blue!50!black, title=Article, 
boxsep=1pt, top=1pt, bottom=1pt, toptitle=1pt, bottomtitle=1pt,
fonttitle=\scriptsize\bfseries]
\scriptsize
Manchester City are keen to sign Anderlecht teenager Evangelos Patoulidis. The 14-year-old playmaker is regarded as one of the best talents to emerge from Anderlecht's youth set-up and has also attracted attention from Arsenal and Barcelona. The Belgian starlet rejected a move to Barcelona's La Masia academy when he was 12 as his family wanted him to continue his studies. He has continued to impress and City have held discussions with Anderlecht chairman Roger Vanden Stock in the hope of agreeing a compensation package. Manuel Pellegrini is looked to build for the future by snapping up hot property Evangelos Patoulidis.
\end{tcolorbox}
\vspace{-12pt}
\begin{tcolorbox}[colback=green!5, colframe=green!50!black, title=Summary, 
boxsep=1pt, top=1pt, bottom=1pt, toptitle=1pt, bottomtitle=1pt,
fonttitle=\scriptsize\bfseries]
\scriptsize
Evangelos Patoulidis is regarded as one of the best players to emerge from Anderlecht youth. He has also attracted attention from Arsenal and Barcelona. The Belgian starlet rejected a move to Barcelona's La Masia academy. The 14-year-old has attracted interest from Barcelona to Barcelona.
\end{tcolorbox}
\vspace{-12pt}
\begin{tcolorbox}[width=\linewidth, colback=red!5, colframe=red!50!black, title=Annotation Scores, 
boxsep=1pt, top=1pt, bottom=1pt, toptitle=1pt, bottomtitle=1pt,
fonttitle=\scriptsize\bfseries]
\scriptsize
\begin{tabular}{ll}
Coherence: & 3 \\
Consistency: & 5 \\
Fluency: & 4 \\
Relevance: & 3 \\
\end{tabular}
\end{tcolorbox}

\caption{Annotating a summary from the SummEval benchmark with scores ranging from 1 to 5 on the metrics: coherence, consistency, fluency and relevance.}
\label{fig:summeval-evaluation-example}
\end{figure}
%%%%% SummEval Example Ends Here

\paragraph{SummEval}
This is a summarization benchmark with 1700 examples where judges rate model-generated summaries of source documents on a 1–5 scale across four metrics: \textbf{coherence} (how well the sentences in the summary fit together), \textbf{consistency} (the factual accuracy of the summary), \textbf{fluency} (grammatical correctness and stylistic quality of each sentence in the summary), and \textbf{relevance} (whether the summary accurately captures the article's main points without including unnecessary details). It is licensed under the MIT license. Each example includes scores from both 3 expert and 5 crowd-sourced annotators. The authors presenting this benchmark reported some agreement metrics. They found the inter-annotator interval kappa to be below an acceptable range, 0.492 and 0.413 for the crowd-sourced workers and the first round of expert annotations accordingly. However, the second round of expert annotations improved the inter-annotator agreement achieving a kappa coefficient of 0.7127.

%%%%% MTBench Example Starts Here
\begin{figure}[ht]
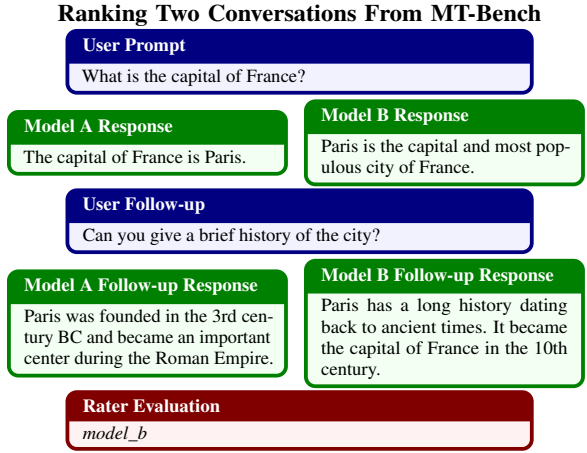

\centering
\begin{tcolorbox}[width=\linewidth, colback=white, colframe=white, halign=center, fontupper=\footnotesize\bfseries, 
boxsep=0pt, top=1pt, bottom=1pt]
Ranking Two Conversations From MT-Bench
\end{tcolorbox}
\vspace{-15pt}
% User initial prompt
\begin{tcolorbox}[width=0.8\linewidth, colback=blue!5, colframe=blue!50!black, title=User Prompt,
boxsep=1pt, top=1pt, bottom=1pt, toptitle=1pt, bottomtitle=1pt,
fonttitle=\scriptsize\bfseries]
\scriptsize
What is the capital of France?
\end{tcolorbox}
\vspace{-6pt}
% Model A and B responses
\noindent
\begin{minipage}{0.48\linewidth}
\begin{tcolorbox}[colback=green!5, colframe=green!50!black, title=Model A Response,
boxsep=1pt, top=1pt, bottom=1pt, toptitle=1pt, bottomtitle=1pt,
fonttitle=\scriptsize\bfseries]
\scriptsize
The capital of France is Paris.
\end{tcolorbox}
\end{minipage}
\hfill
\begin{minipage}{0.48\linewidth}
\begin{tcolorbox}[colback=green!5, colframe=green!50!black, title=Model B Response,
boxsep=1pt, top=1pt, bottom=1pt, toptitle=1pt, bottomtitle=1pt,
fonttitle=\scriptsize\bfseries]
\scriptsize
Paris is the capital and most populous city of France.
\end{tcolorbox}
\end{minipage}
\vspace{-6pt}
% Follow-up question
\begin{tcolorbox}[width=0.8\linewidth, colback=blue!5, colframe=blue!50!black, title=User Follow-up,
boxsep=1pt, top=1pt, bottom=1pt, toptitle=1pt, bottomtitle=1pt,
fonttitle=\scriptsize\bfseries]
\scriptsize
Can you give a brief history of the city?
\end{tcolorbox}
\vspace{-6pt}
% Follow-up responses
\noindent
\begin{minipage}{0.48\linewidth}
\begin{tcolorbox}[colback=green!5, colframe=green!50!black, title=Model A Follow-up Response,
boxsep=1pt, top=1pt, bottom=1pt, toptitle=1pt, bottomtitle=1pt,
fonttitle=\scriptsize\bfseries]
\scriptsize
Paris was founded in the 3rd century BC and became an important center during the Roman Empire.
\end{tcolorbox}
\end{minipage}
\hfill
\begin{minipage}{0.48\linewidth}
\begin{tcolorbox}[colback=green!5, colframe=green!50!black, title=Model B Follow-up Response,
boxsep=1pt, top=1pt, bottom=1pt, toptitle=1pt, bottomtitle=1pt,
fonttitle=\scriptsize\bfseries]
\scriptsize
Paris has a long history dating back to ancient times. It became the capital of France in the 10th century.
\end{tcolorbox}
\end{minipage}
\vspace{-6pt}
% Final judgment
\begin{tcolorbox}[width=0.8\linewidth, colback=red!5, colframe=red!50!black, title=Rater Evaluation,
boxsep=1pt, top=1pt, bottom=1pt, toptitle=1pt, bottomtitle=1pt,
fonttitle=\scriptsize\bfseries]
\scriptsize
\textit{model\_b}
\end{tcolorbox}

\caption{Ranking conversations from MT-bench with by indicating \textit{model\_a}, \textit{model\_b} or \textit{tie}.}
\label{fig:mtbench-evaluation-example}
\end{figure}
%%%%% MTBench Example Ends Here

\paragraph{MT-Bench}
In a multi-turn conversation from this benchmark, a user prompts two LLMs with a question, and after receiving their responses, asks a follow-up question, to which a second pair of responses is generated. A rater (human or LLM) is shown this conversation and has to assign a label (\textit{model\_a}, \textit{model\_b}, \textit{tie}) indicating model preference. Figure~\ref{fig:mtbench-evaluation-example} shows an example of a conversation from this dataset getting rated.

There are 80 questions that require an LLM to perform multi-turn conversations and follow instructions on topics including reasoning and math. There are 30 examples for each question, where each example in the dataset comprises a question followed by responses from two models, \textit{model\_a} and \textit{model\_b}, for a total of 2.4k examples. Each example also contains a judgment assigned by GPT-4 and judgments from zero to five human raters. Since we cannot perform any agreement analysis between human raters for fewer than 2 raters, we create a smaller filtered subset of the data containing 761 examples where each example contains two or more human ratings. Table~\ref{tab:mtbench_count_judges} in Appendix~\ref{app:mtbench_examples} shows the distribution of the number of human ratings in the dataset that we use in our experiments.

\subsection{Experimental Settings}
For all three benchmarks, three Large Language Models (LLMs) are used as judges. We use the following models across all benchmarks: Llama-3.1-70B-Instruct~\citep{llama31}, DeepSeek-R1-Distill-Qwen~\citep{deepseekai2025deepseekr1incentivizingreasoningcapability}, and Qwen3-32B~\citep{yang2025qwen3technicalreport}. Details of the hyperparameter configurations and prompt templates are available in Appendix~\ref{app:experimental_setup}.

For all three benchmarks, we ran each judge LLM on the same set of generations independently for three runs. We used the same prompts and settings for each run to measure intra-rater variance under fixed conditions. For \textbf{SummaC}, we added the articles and the corresponding generated summaries from the test set to be rated by each model to the prompt. Once we had three sets of scores for a benchmark and for an LLM judge we computed intra-rater reliability using Krippendorff's Alpha. In our initial experiments, we also tried running our LLM judges on additional runs (up to 10) but found no significant effect of the number of runs on self-reliability so ultimately, we kept the number of runs to 3.

\textbf{SummEval} allows reliability analysis along specific evaluation metrics: Coherence, Consistency, Fluency, and Relevance. We prompted each LLM judge to evaluate each metric on a scale of 1 to 5 independently per run. That means for each of the three runs we prompted the LLM four times, once for each metric. To better align with ordinal rating behavior, we replace SummEval's default interval-based distance metric with ordinal distance for computing agreement, as the ratings are on a 1–5 Likert scale. This allows a more principled estimate of intra-rater reliability by accounting for ordinal semantics.

For \textbf{MTBench}, we closely follow the setup from~\citet{zheng2023judgingllmasajudgemtbenchchatbot}, with the addition of measuring judge consistency when evaluating chatbot responses over three runs per LLM judge. Each LLM judge scores the same prompts using the same interface scripts as the original benchmark. In addition to intra-rater reliability, we also analyze human-human and human-LLM agreement using the expert and crowd-sourced annotations provided in the benchmark, with human judgments from GPT-4 used as a reference point for LLM-human comparisons.

The prompts used in each of these benchmarks are in Appendix.

\subsection{Agreement Metric}
\label{sec:agreement_metric}
As we have multiple independent ratings from an LLM judge on a single item, we can adapt existing agreement metrics to measure self-reliability of an LLM judge. For both self-reliability of an LLM judge and its agreement with other judges, we primarily use \textbf{Krippendorff’s Alpha}. We describe this metric and the choice of distance functions in Appendix~\ref{app:ka_description}. Additionally, we also use this metric to report inter-rater agreement across categories of judges (e.g., LLM vs. human, or expert vs. crowd).

\paragraph{Why Krippendorff's Alpha} Other works studying LLM-as-a-judge typically use different metrics for computing agreement between LLM and human judges. For example, G-Eval~\citep{liu-etal-2023-g} used correlation and MTBench~\citep{zheng2023judgingllmasajudgemtbenchchatbot} used accuracy. So why go with a different metric? \citet{artstein-poesio-2008-survey} discussed several drawbacks when metrics without chance-correction like accuracy and correlation. Despite being intuitive to understand, their values cannot be compared across studies, because some of the agreement will be due to chance, and that chance is affected by factors that vary from one study to another. One factor is that chance agreement is higher when there are fewer categories or labels. For example, in binary classification with uniformly distributed labels, the chance agreement will be 50\%, whereas when we have three labels it will be 33\%. Another reason these metrics cannot be trusted is that they do not correct for for the distribution of items across categories. For example, in binary classification, if one label appears 95\% of the time, the chance agreement for that label will be $0.95 \times 0.95 = 0.9025$, and it will be $0.05 \times 0.05 = 0.0025$ for the other label. This would mean that two raters would be expected to agree $90.5\%$ of the time, and an observed agreement of $90\%$ may look high but is actually worse than what we would expect to get by chance. A chance-corrected metric like Krippendorff's Alpha addresses these drawbacks. It also has advantages over other chance-corrected metrics like Cohen's Kappa~\citep{cohenkappa} and Fleiss' Kappa~\citep{Fleiss1971} with more flexibility supporting varying numbers of annotators, handling missing data, and accommodating different distance functions suited to the rating scale of each benchmark. Due to these reasons, we believe Krippendorff's Alpha is appropriate for measuring both self-reliability in LLM judges and their agreement with human judges.

We adapt the underlying distance function for Krippendorff’s Alpha based on the nature of the ratings in each benchmark. In \textbf{SummaC}, labels are binary and categorical, so we use \textit{nominal distance}, which treats all disagreements equally without assuming an ordering. We also report \textbf{Balanced Accuracy}—the mean of sensitivity and specificity—when comparing LLM and human judgments, following the benchmark’s original evaluation setup and accounting for class imbalance.

In \textbf{SummEval}, ratings are ordinal, so we use \textit{ordinal distance} to reflect varying degrees of disagreement (e.g., a 1-point difference counts less than a 3-point difference). Agreement is computed between cross-category judge pairs only, excluding intra-category comparisons.

For \textbf{MTBench}, we use two metrics: (1) \textbf{Accuracy}, matching the original benchmark, measures agreement with the gold standard (majority human vote); (2) Krippendorff’s Alpha with \textit{ordinal distance} captures gradations in pairwise disagreement (e.g., \textit{model\_a} vs. \textit{tie} vs. \textit{model\_b}). Both inter-rater (e.g., LLM vs. human) and intra-rater (across LLM runs) agreement are reported.

%%%% Add justification on why Krippendorff's Alpha is better and correlation (which was used in G-Eval) and accuracy (which was used in MTBench) are bad.

\section{Intra-Rater Reliability Results}
\paragraph{SummaC} Table~\ref{tab:self-reliability-all} shows the Krippendorff's Alpha computed for LLM judges over 3 runs. We see that the agreement value is low for all models for all runs, though it gets better for newer and larger models, with Qwen 3 getting close to 0.8, which is the commonly accepted threshold of good agreement. We tried repeating these experiments for up to 5 runs but found no significant difference in self-reliability, suggesting that this value is fairly static for a model on a specific test set, independent of the number of runs.

\begin{table}[]
\centering
\small
\begin{tabular}{@{}ll@{}}
\toprule
Model       & Self-Reliability \\ \midrule
Llama 3.1   & 0.3263           \\
Deepseek-R1 & 0.6278           \\
Qwen-3      & 0.7883           \\ \bottomrule
\end{tabular}
\caption{Self-reliability of different LLM judges on the SummaC benchmark across 3 runs measured by Krippendorff's Alpha.}
\label{tab:self-reliability-all}
\end{table}

\begin{figure}[h]
\centering
\includegraphics[width=\columnwidth]{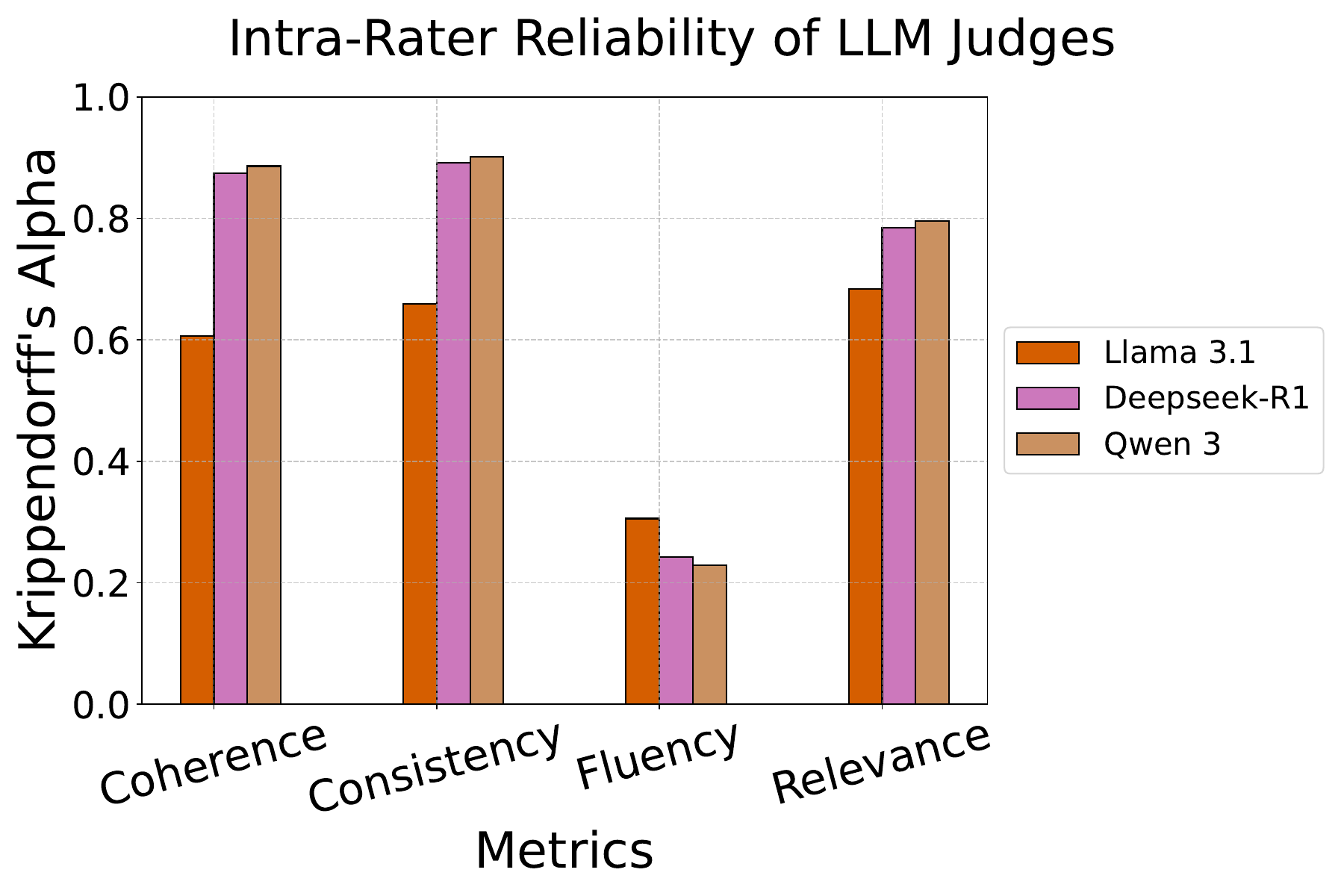}
\caption{Self-reliability of LLM judges on SummEval.}
\label{fig:pairwise_llm_agreement}
\end{figure}

\paragraph{SummEval} In Figure~\ref{fig:pairwise_llm_agreement}, we see that compared to human evaluators, the different runs in Deepseek-R1 and Qwen 3 show a high self-reliability on Coherence and Consistency, with very low self-reliability on Fluency. Not only does Fluency have low reliability for all our LLM judges, but it is also the only metric where Llama performs best. This suggests that an LLM judge's ability to reliably produce ratings depends heavily on the metric.

\paragraph{MTBench} We found that the self-reliability of LLM raters is much lower on this benchmark compared to SummaC and SummEval. Llama 3.1 had a Krippendorff's Alpha of \textbf{0.265} across its 3 runs, while for Deepseek-R1 it was \textbf{0.507}. Qwen 3 was the most reliable with a Krippendorff's Alpha of \textbf{0.563}. Even in the best case, these numbers are much lower than the desired threshold of 0.8, showing that LLM raters are extremely volatile on this task, even more so than summarization. In fact, Qwen 3 gave the same judgment on all 3 runs for only 61.3\% of cases.

\section{Are LLM Judges a Reliable Substitute for Human Judges?}
Our next set of experiments studies what this lack of self-reliability implies in terms of real-world performance, which for benchmarks usually means comparing with human annotations. In \textbf{SummaC}, we measure the agreement of LLM judgments with human annotations, while in \textbf{SummEval} and \textbf{MTBench}, we have multiple human labels available per example, allowing us to compare the agreement between LLM judges and human judges with human-human agreement.

\subsection{SummaC}
\label{sec:summac_agreement_results}

\begin{figure}[h]
\centering
\includegraphics[width=\columnwidth]{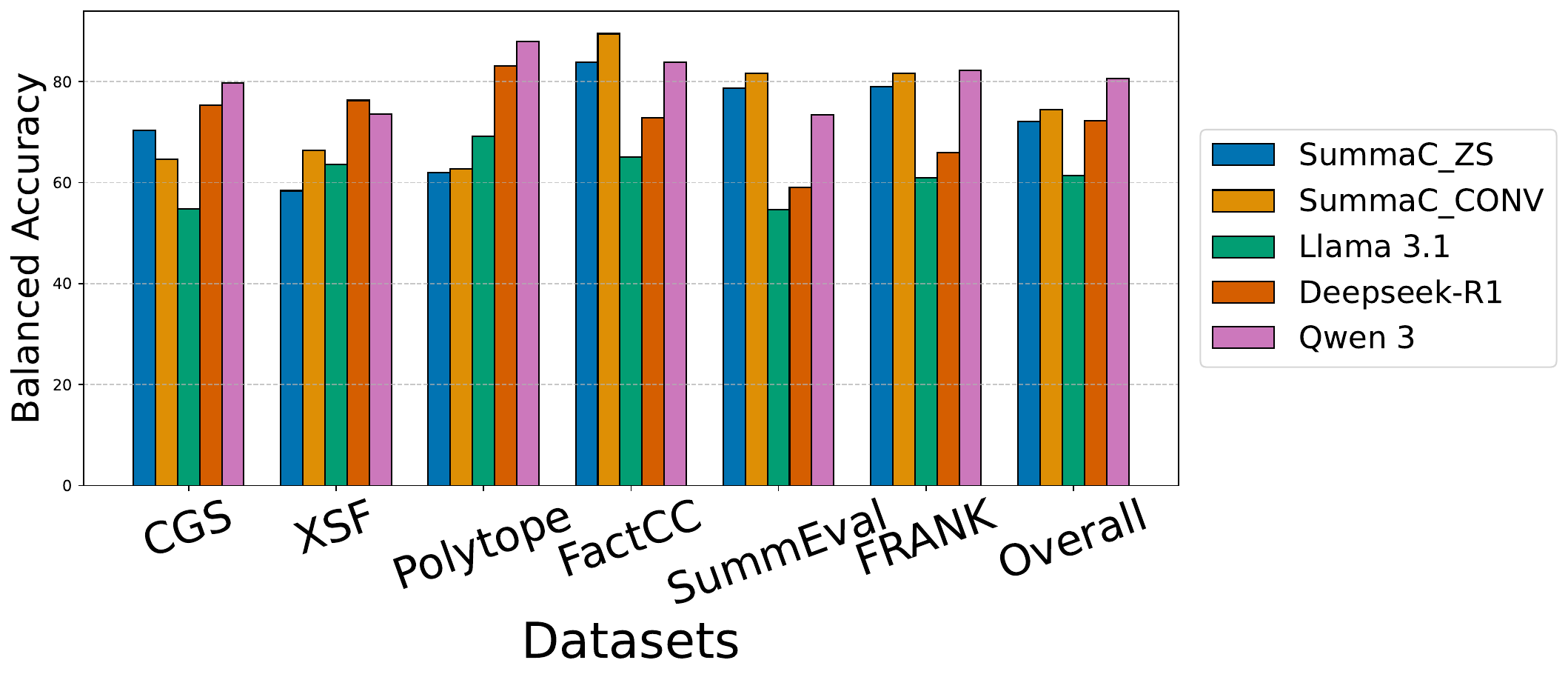}
\caption{Balanced Accuracies of different LLM judges against human judgments on the datasets in SummaC.}
\label{fig:summac_dataset_breakdown}
\end{figure}

In Figure~\ref{fig:summac_dataset_breakdown}, we report the balanced accuracies of our three LLM judges on the different datasets within SummaC, along with the two baselines, $\mathit{SummaC_{ZS}}$ and $\mathit{SummaC_{CONV}}$ that were introduced by~\citet{laban-etal-2022-summac}. Since we had 3 runs for each LLM judge, we compute the consensus rating between the 3 runs by taking a simple majority vote and then find the agreement between it and the human label. We see that while Llama 3.1 has very low performance, Deepseek-R1 is competitive with the baselines and Qwen 3 significantly outperforms the baseline models overall. It is to be noted that the baselines $\mathit{SummaC_{ZS}}$ and $\mathit{SummaC_{CONV}}$ were fine-tuned on this task whereas in our experiments we prompted the instruction-tuned models off-the-shelf.

Despite performing best overall, Qwen 3 is not the best at all datasets in the benchmark. Deepseek-R1 performs the best at XSumFaith. $\mathit{SummaC_{CONV}}$ performs the best at FactCC, which is unsurprising since that baseline was specifically fine-tuned on training examples from that dataset~\citep{laban-etal-2022-summac}. Therefore, how well an LLM can substitute a human judge varies greatly on the dataset in question, though in general, newer and larger models perform better.

In Table~\ref{tab:acc_with_humans}, we see that accuracy varies between runs much more for Llama 3.1, while it is relatively stable for Deepseek-R1 and Qwen 3.

\begin{table}[t]
\centering
\resizebox{\columnwidth}{!}{%
\begin{tabular}{@{}llll@{}}
\toprule
Model       & Single Run (Mean $\pm$ Std)    & Majority & No Sampling \\ \midrule
\textit{Llama 3.1}   & 59.1 $\pm$ 2.06 & 61.4   & 58.4       \\
\textit{Deepseek-R1} & 69.8 $\pm$ 0.50 & 72.3   & 69.3       \\
\textit{Qwen 3} & 79.4 $\pm$ 0.32 & 80.6   & 79.2       \\
\bottomrule
\end{tabular}
}
\caption{Balanced accuracies of LLM judgments against human judgments under different experimental settings.}
\label{tab:acc_with_humans}
\end{table}

\paragraph{Does Taking Majority Vote Help Performance?} In Table~\ref{tab:acc_with_humans}, for each LLM, we report the balanced accuracy of a single run, the accuracy we get if we take the consensus rating of the three runs instead, as well as the accuracy if we turn off sampling to ensure we get the same output every time. We see that accuracy is higher when computed via majority vote than the expected accuracy for a single run. In fact, for Deepseek-R1 and Qwen 3, taking the majority vote gives a higher balanced accuracy than the maximum accuracy achieved by a single run. Not only does running the LLM multiple runs accounts for this variance between runs, but it also gives higher performance in terms of accuracy against human judgments. This performance is also competitive with the baselines $\mathit{SummaC_{ZS}}$ and $\mathit{SummaC_{CONV}}$~\citep{laban-etal-2022-summac}, which were fine-tuned specifically for this task.

\paragraph{Does disabling temperature sampling reduce variance without degrading performance?}
The main reason this issue exists is that these LLMs are designed to perform sampling to generate slightly different responses to the same prompt. Therefore, to remove variance in ratings, the most obvious answer is to run inference without any sampling to get the same output every time. However, in Table~\ref{tab:acc_with_humans}, we see that for both models, there is a degradation in performance if run without sampling. This shows that there is a trade-off between self-reliability and performance, and it is not trivial to address both issues simultaneously. We also ran additional experiments in Appendix~\ref{app:summac_prompting} to see if incorporating few-shot or chain-of-thought prompting led to any reliability or agreement gains and found that to not be the case.

% \section{Agreement Between Human Judges}

\begin{figure}[h]
\centering
\includegraphics[width=\columnwidth]{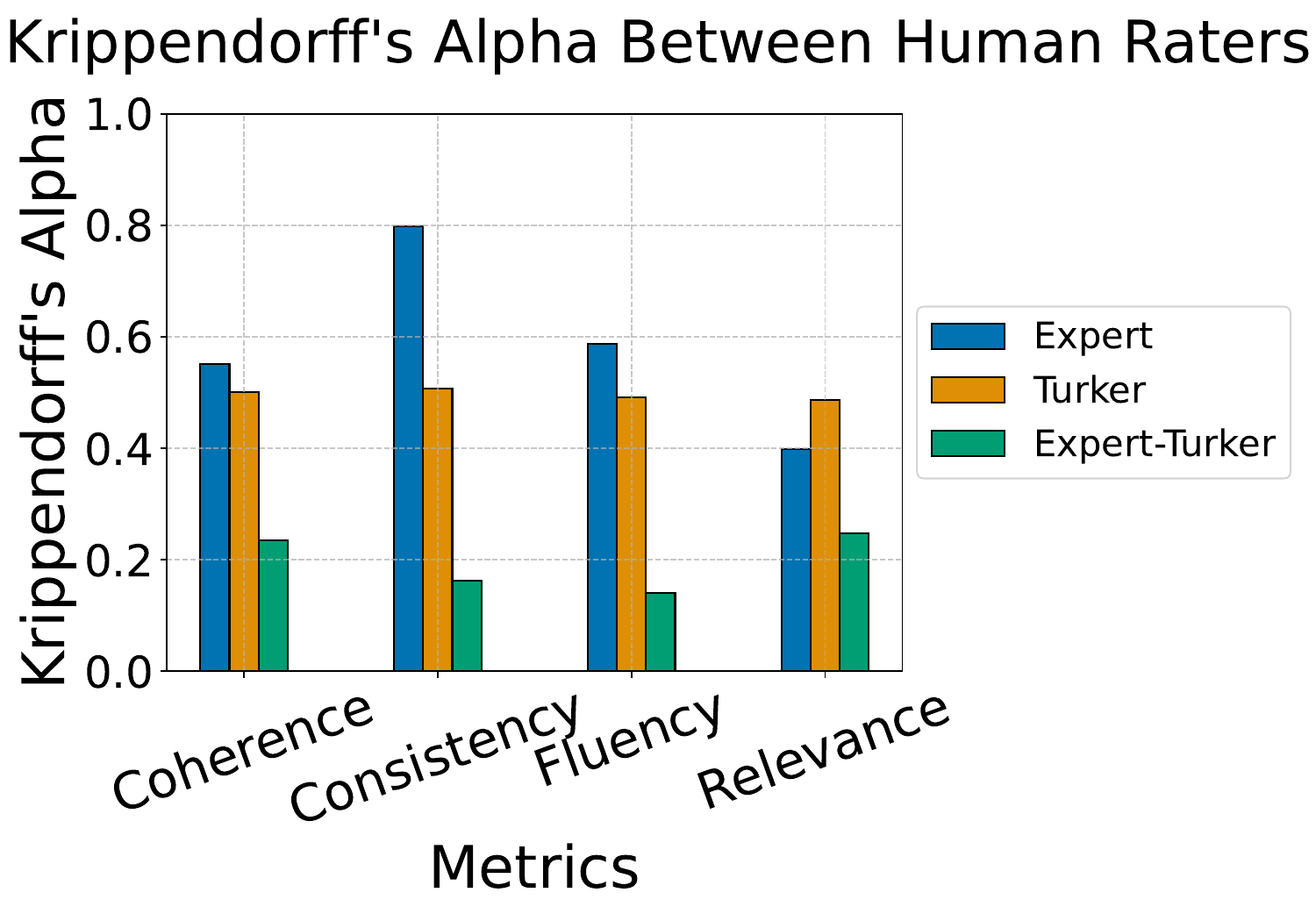}
\caption{Inter-rater reliability within and across both categories of human judges on SummEval.}
\label{fig:pairwise_human_agreement}
\end{figure}

\subsection{SummEval} 
\label{sec:summeval_agreement_humans}Figure~\ref{fig:pairwise_human_agreement} presents Krippendorff's Alpha agreement scores among human annotators. Experts show the highest agreement on consistency (0.798), moderate agreement on fluency (0.588), and the lowest on relevance (0.398), suggesting subjectivity in assessing relevance. In contrast, crowdworkers (Turkers) display uniformly lower agreement across all metrics (0.48–0.51), indicating more limited task comprehension. Agreement between experts and Turkers is drastically lower (maximum 0.247 for relevance), highlighting fundamental differences in the evaluation approaches of different populations of raters.

\begin{figure}[h]
\centering
\includegraphics[width=\columnwidth]{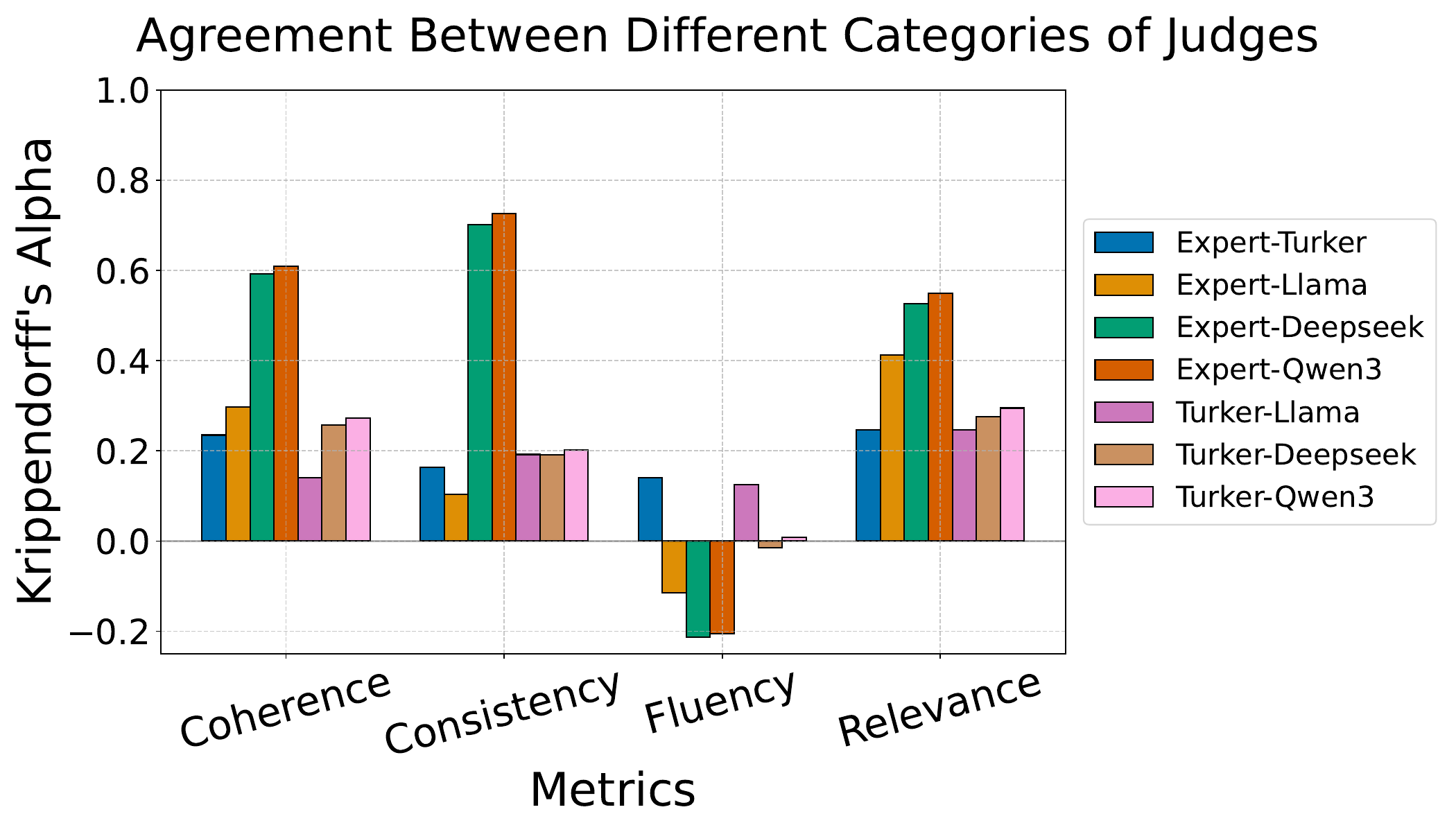}
\caption{Inter-rater reliability of LLM judges against both types of human judgments on SummEval.}
\label{fig:pairwise_inter_agreement}
\end{figure}

While LLM judges show higher agreement with themselves than human judges do with each other, this does not guarantee they can fully replace human judges. In Figure~\ref{fig:pairwise_inter_agreement}, we see that LLMs’ agreement with human experts varies significantly depending on the metric, dataset, and the expertise of the human judge.

For instance, Turker ratings diverge from both experts and LLMs, indicating they use conflicting evaluation criteria. Experts and LLMs show somewhat higher, but still modest agreement, especially on metrics like consistency and coherence, while agreement drops or even turns negative for subjective metrics such as fluency.

Model scaling (e.g., using larger models like Qwen 3) can improve agreement with experts for some metrics, but even the best observed agreement (0.726 for consistency) remains below the commonly accepted threshold for substitution. Moreover, score distributions differ across judge types and metrics, as reported in Appendix~\ref{app:score_dist}, further complicating direct replacement.

\subsection{MTBench}
We observe a fairly low Krippendorff's Alpha of \textbf{0.478} between human judges. This follows from our results under SummEval, where we find that crowdsourced judges tend to have very lower agreement. However, accuracy is comparatively much higher at \textbf{0.827}, suggesting that it may be a misleading metric when showing agreement between multiple judges, since it does not take the probability of chance agreement into account, leading to inflated scores. A complete breakdown of agreements of different judges against humans is shown in Table~\ref{tab:mtb_agreement}.

Table~\ref{tab:mtb_agreement} shows the agreement between different judge types on the MT-bench benchmark.

\begin{table}[h!]
\centering
% \small
\resizebox{\columnwidth}{!}{%
\begin{tabular}{@{}lll@{}}
\toprule
Judges (Human vs.)             & Accuracy & Krippendorff's Alpha \\ \midrule
Human                          & 0.827             & 0.478                \\
GPT-4                          & 0.671             & 0.396                \\
Llama 3.1      & 0.556             & 0.239                \\
Deepseek-R1    & 0.668             & 0.385                \\
Qwen 3    & 0.719             & 0.426                \\
\bottomrule
\end{tabular}
}
\caption{Agreement of different judges against humans.}
\label{tab:mtb_agreement}
\end{table}

\paragraph{Agreement between LLMs and humans}
Similar to agreement between humans, we see that when we use Krippendorff's alpha, we get much lower values of agreement than accuracy. We see that GPT-4 has higher values of agreement across both metrics, though it is still much lower than between humans. However, even for humans this is far lower than the accepted threshold of 0.8, and is much lower than the agreement we saw for SummEval, suggesting this task is even harder to consistently than summarization.

\section{Discussion}
\subsection{Takeaways}
\paragraph{SummaC}
Even though simply prompting an LLM to score a generation can perform comparable to models specifically fine-tuned for this task~\citep{laban-etal-2022-summac}, it calls into question what high performance means in this context when there is high variance between runs. For example, what does an agreement of 0.8 between LLMs and human judges mean when the LLM has low agreement with itself? We also do not know if this phenomenon was due to the limitations of the benchmark itself (e.g. noisy examples or a poorly defined task) or is more widespread across other benchmarks or tasks. More importantly, we do not know if human judgments, which are taken as the gold standard, suffer from the same consistency issues as LLMs, which should prompt further investigations. However, SummaC contains only one human judgment per example, making such investigations impossible, so we must look to other benchmarks and tasks. We saw the LLM judges struggle particularly with one of the datasets within this benchmark, SummEval~\citep{summeval}. In this dataset, we also have multiple human judgments for each example across multiple metrics, which would allow us to address a lot of these questions that so far remained unanswered.

\paragraph{SummEval}
We see that the problem of low intra-rater reliability of LLM judges is not just due to binary labels in the SummaC benchmark but also persists across other metrics and scoring scales. Not only does it make LLM judges unreliable, but it also makes meta-evaluation of different evaluation frameworks difficult when the scores assigned by the same judge fluctuate over time. This raises another important question, is the low intra-rater reliability due to limitations of the LLMs or is the task of summarization not well-defined even with different metrics? It is possible that scoring the summaries in the SummEval benchmark along metrics like coherence or consistency is not a well-defined task, which causes the scores to fluctuate for both human and LLM judges. We need to study whether this phenomenon persists for benchmarks in other tasks.

\paragraph{MT-Bench}
This confirms that it is not just single-turn tasks like summarization where intra-rater reliability is an issue. It is a widespread phenomenon across benchmarks as well as tasks that involve text generation. We also observe that metrics like accuracy inflate agreement values compared to metrics like Krippendorff's Alpha.

\subsection{Recommendations}
\label{sec:recommendations}
Based on these findings, we can make the following recommendations to improve the state of NLG evaluation with LLM as well as human judges.

\paragraph{Account For Possible Self-Reliability Issues} Future work should incorporate intra-rater reliability information into evaluation frameworks and explore methods to improve LLM self-reliability without sacrificing agreement with human judgments. Existing NLG research already uses metrics like \textbf{Cohen's Kappa} and \textbf{Krippendorff's Alpha} for measuring inter-rater reliability, and these metrics can be adapted to measure self-reliability as well. They should be preferred over metrics like accuracy, since as we saw in Table~\ref{tab:mtb_agreement} that accuracy inflates agreement numbers because it does not take chance agreement into account.

\paragraph{Reduce Variance in LLM Outputs} We saw in Section~\ref{sec:summac_agreement_results} that taking an aggregate of the results of multiple runs, like a simple majority vote, can improve agreement with human judgment. On the other hand, trying to eliminate variance entirely by turning off variance hurts performance.

\paragraph{Collect Data on Self-Reliability of Human Judges} When collecting human evaluations on NLG tasks, it is important to also measure self-reliability data on those evaluations. This is because although we saw self-reliability is a problem in LLM judges, we did not explore how prevalent this problem is in human judges. As we consider human judgments the gold standard in NLG evaluation, the self-reliability of human judges would serve as an upper bound of the self-reliability we should reasonably expect from LLM judges. It is also important to explore how much of an effect training or expertise has on self-reliability, since we saw in Section~\ref{sec:summeval_agreement_humans} that expert and crowdsourced workers assign very different ratings.

\section{Conclusion}
We present a comprehensive analysis of intra-rater reliability in LLM-as-a-judge frameworks, revealing key challenges for evaluation in this domain. Our experiments show that LLMs display low self-reliability when evaluating the same content across multiple runs, even with identical prompts and hyperparameters. This inconsistency persists across various tasks and metrics. Although newer, larger models like Qwen 3 are more consistent than models such as Llama 3.1, they still often fall short of standard reliability thresholds.

Our work carries some potential risks. We have measured the performance of LLM judges by their agreement with human judges, which can be noisy depending on the inter-annotator agreement of the human labels. LLM judgments could also potentially have biases for certain linguistic styles which could influence the scores presented in this paper. Despite these risks, our findings have important implications for LLM-as-a-judge research. Reporting single-run LLM judgments without consistency metrics can be misleading, and certain aspects of text quality, such as fluency, remain difficult to assess reliably for both humans and machines. Also, the pronounced disagreement between expert and crowdsourced judges highlights the need to clarify which human preferences LLMs should model.

While LLM-as-a-judge approaches offer scalability, they face significant reliability challenges. Addressing these issues is essential for developing evaluation frameworks that meaningfully capture distinctions in text quality.

\section*{Limitations}
Recent research has looked at analyzing reasoning traces of LLM judges~\citep{wang2023selfconsistencyimproveschainthought}, whether these reasoning traces are useful and how their performance is affected by supervised fine-tuning or reinforcement learning~\citep{chen2025judgelrmlargereasoningmodels}. While we leveraged reasoning models like Deepseek R1 and Qwen 3 in our experiments, we have not explored any relationships between their reasoning traces and their self-reliability. And while prompting LLM judges on benchmarks like SummaC surpasses baselines in terms of agreement with human judgments, we have not looked at the effect of fine-tuning on further improving performance as that is beyond the scope of the paper. Beyond fine-tuning, there are other potential avenues worth exploring like investigating the impact of prompt structure on reliability and applying probing the model internals to study if a specific layer leads to the divergence in assigned scores in conflicting runs. Finally, though we identified that the subjectivity of summarization or multi-turn conversation evaluation potentially exacerbates the self-reliability issue of LLM judges, we have not quantified this subjectivity through comparison with self-reliability on other, more objective tasks. Nevertheless, this work highlights an important issue in evaluating generation tasks and discusses different experimental methods and data collection principles that would allow us to conduct more robust evaluations.

Finally, we limited our choice of models to those with open weights, as this gave us the freedom to run our experiments locally and with full access to the model weights. However, the latest proprietary models such as GPT-5~\citep{openai2025gpt5} and Claude-4~\citep{anthropic2025claude4} have made great strides in various reasoning and agentic benchmarks. Along with the aforementioned future directions suggested, a comparison of the self-reliability in these models with the models in our study would give additional insights into the prevalence of this phenomenon.

\section*{Acknowledgments}
This research used the Delta advanced computing and data resource which is supported by the National Science Foundation (award OAC 2005572) and the State of Illinois. Delta is a joint effort of the University of Illinois Urbana-Champaign and its National Center for Supercomputing Applications.

% This document has been adapted
% by Steven Bethard, Ryan Cotterell and Rui Yan
% from the instructions for earlier ACL and NAACL proceedings, including those for
% ACL 2019 by Douwe Kiela and Ivan Vuli\'{c},
% NAACL 2019 by Stephanie Lukin and Alla Roskovskaya,
% ACL 2018 by Shay Cohen, Kevin Gimpel, and Wei Lu,
% NAACL 2018 by Margaret Mitchell and Stephanie Lukin,
% Bib\TeX{} suggestions for (NA)ACL 2017/2018 from Jason Eisner,
% ACL 2017 by Dan Gildea and Min-Yen Kan,
% NAACL 2017 by Margaret Mitchell,
% ACL 2012 by Maggie Li and Michael White,
% ACL 2010 by Jing-Shin Chang and Philipp Koehn,
% ACL 2008 by Johanna D. Moore, Simone Teufel, James Allan, and Sadaoki Furui,
% ACL 2005 by Hwee Tou Ng and Kemal Oflazer,
% ACL 2002 by Eugene Charniak and Dekang Lin,
% and earlier ACL and EACL formats written by several people, including
% John Chen, Henry S. Thompson and Donald Walker.
% Additional elements were taken from the formatting instructions of the \emph{International Joint Conference on Artificial Intelligence} and the \emph{Conference on Computer Vision and Pattern Recognition}.

% Bibliography entries for the entire Anthology, followed by custom entries
\bibliography{anthology,custom}
% Custom bibliography entries only
% \bibliography{custom}

\appendix
\section{More Details About Experimental Setup}
\label{app:experimental_setup}
For all three benchmarks and all three LLMs in our experiments, we used a 4xA100 GPU server and the transformers library~\citep{wolf2020huggingfacestransformersstateoftheartnatural} to run our experiments. Unless specified otherwise, we used the recommended defaults for each model. This means that for Llama 3.1 and DeepSeek-R1, we used a temperature of 0.6 and top\_p of 0.9. For Qwen 3, we used a temperature of 0.6 and top\_p of 0.95.

%----- SummaC Prompt
For the \textbf{SummaC} benchmark, we used the following prompt to get a binary label representing whether the article is consistent or inconsistent. Figure~\ref{fig:summac-prompt} shows the prompt used for evaluating items in this benchmark.

\begin{figure}[htbp]
\centering
\begin{tcolorbox}[
  colback=gray!5!white, 
  colframe=gray!75!black, 
  boxrule=0.4pt,
  sharp corners=south, 
  left=6pt, right=6pt, top=6pt, bottom=6pt,
  width=\linewidth, % ensure full text width
]

\setlist{nosep, leftmargin=1.8em}

\noindent\textbf{Task}: Analyze the summary for factual inconsistencies against the source document. Inconsistencies can be due to:
\begin{enumerate}
    \item \textbf{Hallucinations}: Information added not in the source.
    \item \textbf{Contradictions}: Statements opposing source content.
    \item \textbf{Entity Errors}: Incorrect names/roles/locations.
    \item \textbf{Omissions}: Key points missing from the summary.
    \item \textbf{Temporal Errors}: Wrong sequence/timeframe of events.
\end{enumerate}

\noindent\textbf{Output}: A single number \textbf{0} for consistent summary and \textbf{1} for inconsistent summary.

\noindent\textbf{Document:} \{\{Full source text\}\}

\noindent\textbf{Summary:} \{\{Generated Summary\}\}

\end{tcolorbox}
\caption{Prompt used for each run in SummaC benchmark.}
\label{fig:summac-prompt}
\end{figure}

\begin{figure*}[htbp]
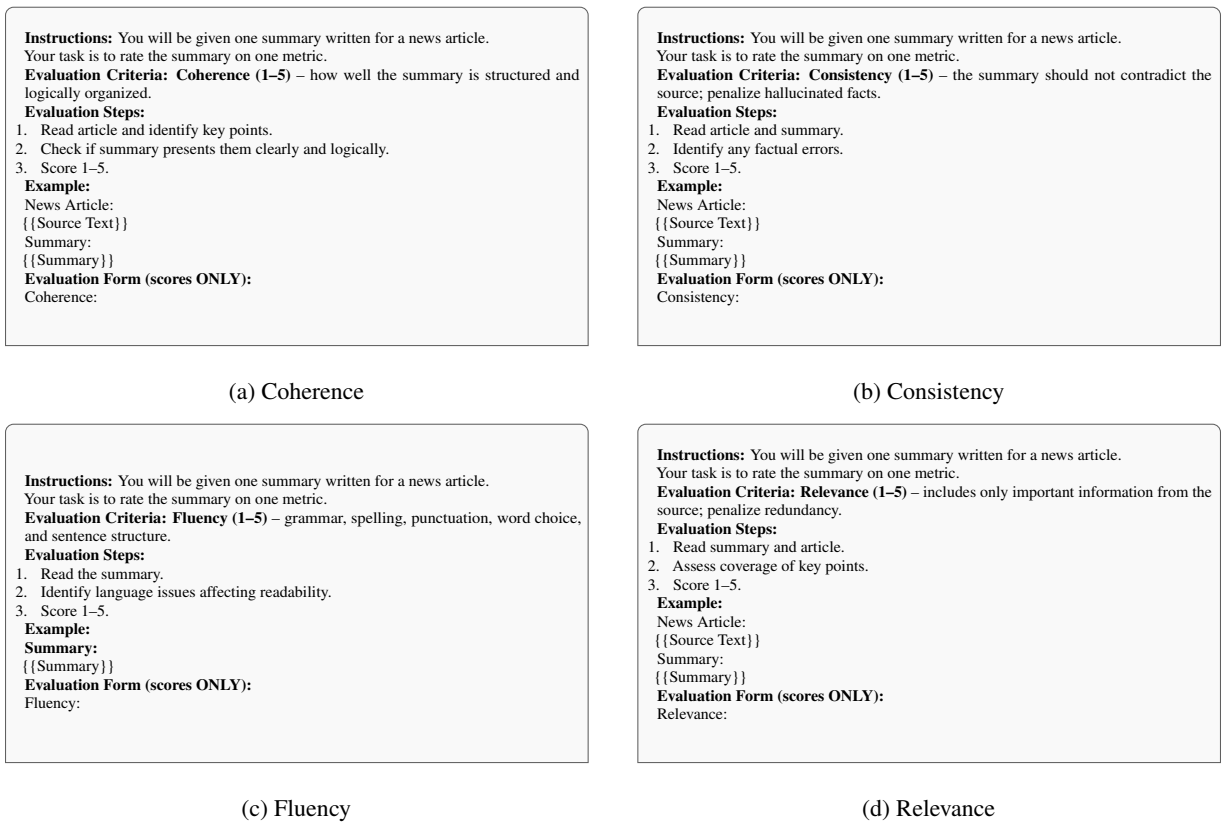

\centering

% 1st row
\begin{subfigure}[t]{0.48\textwidth}
\begin{tcolorbox}[colback=gray!5!white, colframe=gray!75!black, boxrule=0.4pt,
  sharp corners=south,
  left=6pt, right=2pt, top=4pt, bottom=4pt,
  fontupper=\tiny,
  height=4.5cm,
  valign=center,
]
\textbf{Instructions:} You will be given one summary written for a news article.

Your task is to rate the summary on one metric.

\textbf{Evaluation Criteria:} \textbf{Coherence (1–5)} – how well the summary is structured and logically organized.

\textbf{Evaluation Steps:}
\begin{enumerate}[nosep, leftmargin=1em]
  \item Read article and identify key points.
  \item Check if summary presents them clearly and logically.
  \item Score 1–5.
\end{enumerate}

\textbf{Example:} \\
News Article: \\
\{\{Source Text\}\} \\
Summary: \\
\{\{Summary\}\} \\
\textbf{Evaluation Form (scores ONLY):} \\
Coherence: \\
\end{tcolorbox}
\caption{Coherence}
\end{subfigure}
\hfill
\begin{subfigure}[t]{0.48\textwidth}
\begin{tcolorbox}[colback=gray!5!white, colframe=gray!75!black, boxrule=0.4pt,
  sharp corners=south,
  left=6pt, right=2pt, top=4pt, bottom=4pt,
  fontupper=\tiny,
  height=4.5cm,
  valign=center,
]
\textbf{Instructions:} You will be given one summary written for a news article.

Your task is to rate the summary on one metric.

\textbf{Evaluation Criteria:} \textbf{Consistency (1–5)} – the summary should not contradict the source; penalize hallucinated facts.

\textbf{Evaluation Steps:}
\begin{enumerate}[nosep, leftmargin=1em]
  \item Read article and summary.
  \item Identify any factual errors.
  \item Score 1–5.
\end{enumerate}

\textbf{Example:} \\
News Article: \\
\{\{Source Text\}\} \\
Summary: \\
\{\{Summary\}\} \\
\textbf{Evaluation Form (scores ONLY):} \\
Consistency: \\
\end{tcolorbox}
\caption{Consistency}
\end{subfigure}

\vspace{0.5em}

% 2nd row
\begin{subfigure}[t]{0.48\textwidth}
\begin{tcolorbox}[colback=gray!5!white, colframe=gray!75!black, boxrule=0.4pt,
  sharp corners=south,
  left=6pt, right=2pt, top=4pt, bottom=4pt,
  fontupper=\tiny,
  height=4.5cm,
  valign=center,
]
\textbf{Instructions:} You will be given one summary written for a news article.

Your task is to rate the summary on one metric.

\textbf{Evaluation Criteria:} \textbf{Fluency (1–5)} – grammar, spelling, punctuation, word choice, and sentence structure.

\textbf{Evaluation Steps:}
\begin{enumerate}[nosep, leftmargin=1em]
  \item Read the summary.
  \item Identify language issues affecting readability.
  \item Score 1–5.
\end{enumerate}

\textbf{Example:} 

\textbf{Summary:} \\
\{\{Summary\}\}

\textbf{Evaluation Form (scores ONLY):} \\
Fluency:
\end{tcolorbox}
\caption{Fluency}
\end{subfigure}
\hfill
\begin{subfigure}[t]{0.48\textwidth}
\begin{tcolorbox}[colback=gray!5!white, colframe=gray!75!black, boxrule=0.4pt,
  sharp corners=south,
  left=6pt, right=2pt, top=4pt, bottom=4pt,
  fontupper=\tiny,
  height=4.5cm,
  valign=center,
]
\textbf{Instructions:} You will be given one summary written for a news article.

Your task is to rate the summary on one metric.

\textbf{Evaluation Criteria:} \textbf{Relevance (1–5)} – includes only important information from the source; penalize redundancy.

\textbf{Evaluation Steps:}
\begin{enumerate}[nosep, leftmargin=1em]
  \item Read summary and article.
  \item Assess coverage of key points.
  \item Score 1–5.
\end{enumerate}

\textbf{Example:} \\
News Article: \\
\{\{Source Text\}\} \\
Summary: \\
\{\{Summary\}\} \\
\textbf{Evaluation Form (scores ONLY):} \\
Relevance: \\
\end{tcolorbox}
\caption{Relevance}
\end{subfigure}

\caption{Prompts For Evaluating Generated Summaries From SummEval Using Four Metrics}
\label{fig:summeval_prompts}
\end{figure*}

%----- SummEval Prompt
For \textbf{SummEval}, there are four different metrics, coherence, consistency, fluency and relevance. For each run, we prompted the LLM judge four times independently, once for each metric. The prompts are in Figure~\ref{fig:summeval_prompts}.

%----- MTBench Prompt
For MTBench, we used a single prompt that asked an LLM judge to choose between two generations, or assign the label \textit{tie} if it was unable to choose a clear favorite. Items in this datasets are divided into two types, \textit{math} and \textit{general}, with the difference being that conversations labeled \textit{math} also contain reference answers to the user query. Figure~\ref{fig:prompt_mtbench_math} shows the prompt used for items labeled \textit{math}, while Figure~\ref{fig:prompt_mtbench_general} shows the prompt used for items labeled \textit{general}.

\begin{figure*}[htbp]
\centering
\begin{tcolorbox}[colback=gray!5!white, colframe=gray!75!black,
  boxrule=0.4pt,
  sharp corners=south,
  left=6pt, right=2pt, top=4pt, bottom=4pt,
  fontupper=\tiny
]
Please act as an impartial judge and evaluate the quality of the responses provided by two AI assistants to the user questions. Your evaluation should consider correctness and helpfulness. You will be given reference answers, the assistant A's answers, the assistant B's answers. Your job is to determine which assistant provides correct and helpful answers to the second user question. Begin your evaluation by comparing both assistants' answers with the reference answers. Identify and correct any mistakes. Avoid any position biases and ensure that the order in which the responses were presented does not influence your decision. Do not allow the length of the responses to influence your evaluation. Do not favor certain names of the assistants. Be as objective as possible. After providing your explanation, output your final verdict by strictly following this format: "[[A]]" if assistant A is better, "[[B]]" if assistant B is better, and "[[C]]" for a tie.

\medskip

\texttt{<|The Start of Reference Answer|>}

\medskip

\textbf{User:}\\
\{\{question\_1\}\}

\textbf{Reference answer:}\\
\{\{ref\_answer\_1\}\}

\textbf{User:}\\
\{\{question\_2\}\}

\textbf{Reference answer:}\\
\{\{ref\_answer\_2\}\}

\medskip

\texttt{<|The End of Reference Answer|>}

\medskip

\texttt{<|The Start of Assistant A's Conversation with User|>}

\medskip

\textbf{User:}\\
\{\{question\_1\}\}

\textbf{Assistant A:}\\
\{\{answer\_a\_1\}\}

\textbf{User:}\\
\{\{question\_2\}\}

\textbf{Assistant A:}\\
\{\{answer\_a\_2\}\}

\medskip

\texttt{<|The End of Assistant A's Conversation with User|>}

\medskip

\texttt{<|The Start of Assistant B's Conversation with User|>}

\medskip

\textbf{User:}\\
\{\{question\_1\}\}

\textbf{Assistant B:}\\
\{\{answer\_b\_1\}\}

\textbf{User:}\\
\{\{question\_2\}\}

\textbf{Assistant B:}\\
\{\{answer\_b\_2\}\}

\medskip

\texttt{<|The End of Assistant B's Conversation with User|>}

\end{tcolorbox}
\caption{Prompt for Evaluating Two Generated Responses to Queries Labeled \textit{math} in MTBench Dataset}
\label{fig:prompt_mtbench_math}
\end{figure*}

\begin{figure*}[htbp]
\centering
\begin{tcolorbox}[colback=gray!5!white, colframe=gray!75!black,
  boxrule=0.4pt,
  sharp corners=south,
  left=6pt, right=2pt, top=4pt, bottom=4pt,
  fontupper=\tiny
]
Please act as an impartial judge and evaluate the quality of the responses provided by two AI assistants to the user questions. You should choose the assistant that follows the user's instructions and answers the user's questions better. Your evaluation should consider factors such as the helpfulness, relevance, accuracy, depth, creativity, and level of detail of their responses. You should focus on who provides a better answer to the second user question. Begin your evaluation by comparing the responses of the two assistants and provide a short explanation. Avoid any position biases and ensure that the order in which the responses were presented does not influence your decision. Do not allow the length of the responses to influence your evaluation. Do not favor certain names of the assistants. Be as objective as possible. After providing your explanation, output your final verdict by strictly following this format: "[[A]]" if assistant A is better, "[[B]]" if assistant B is better, and "[[C]]" for a tie.

\medskip

\texttt{<|The Start of Assistant A's Conversation with User|>}

\medskip

\textbf{User:}\\
\{\{question\_1\}\}

\textbf{Assistant A:}\\
\{\{answer\_a\_1\}\}

\textbf{User:}\\
\{\{question\_2\}\}

\textbf{Assistant A:}\\
\{\{answer\_a\_2\}\}

\medskip

\texttt{<|The End of Assistant A's Conversation with User|>}

\medskip

\texttt{<|The Start of Assistant B's Conversation with User|>}

\medskip

\textbf{User:}\\
\{\{question\_1\}\}

\textbf{Assistant B:}\\
\{\{answer\_b\_1\}\}

\textbf{User:}\\
\{\{question\_2\}\}

\textbf{Assistant B:}\\
\{\{answer\_b\_2\}\}

\medskip

\texttt{<|The End of Assistant B's Conversation with User|>}

\end{tcolorbox}
\caption{Prompt for Evaluating Two Generated Responses to Queries Labeled \textit{general} in MTBench Dataset}
\label{fig:prompt_mtbench_general}
\end{figure*}

\section{Agreement Metric: Krippendorff's Alpha}
\label{app:ka_description}
Krippendorff’s alpha ($\alpha$) is a chance‐corrected reliability coefficient used to quantify agreement among two or more coders (observers, raters) assigning values to a set of units. In other words, it measures how consistently multiple annotators code the same items. Developed by Klaus Krippendorff in the context of content analysis, $\alpha$ generalizes many classical agreement statistics (such as Cohen’s~\citep{cohenkappa} or Fleiss’~\citep{Fleiss1971} Kappa) and applies to any number of coders, any number of categories or scale values, and any measurement level (nominal, ordinal, interval, ratio, etc.)

Krippendorff’s alpha accommodates any number of raters (two or more) and any measurement level, from nominal categories up through interval (or ratio) scales. It can handle missing data (by simply omitting those cases when counting pairwise judgments), and it yields comparable reliability coefficients even for unequal sample sizes or many categories. This flexibility allows it to remain valid for more than two raters or for ordered categories where partial agreements matter. Overall, Krippendorff’s alpha is valued for its broad applicability and its principled treatment of chance agreement, making it suitable for diverse rating and annotation tasks.

\subsection{Mathematical Notation}

Mathematically, Krippendorff's Alpha follows the general form:

\begin{equation}
    \alpha = 1 - \frac{D_o}{D_e}
\end{equation}

Here, $D_o = \sum \delta(v_i,v_j)$ sums the pairwise distances between raters’ values $v_i$, $v_j$ on each item. 

The denominator $D_e$ is computed by summing $\delta$ over all possible pairs of values (weighted by the frequency of each value) as if the assignments were randomly permuted among raters. In other words, $D_e$ reflects the average disagreement expected purely by chance given the observed distribution of categories. The expected disagreement $D_e$ is what separates Krippendorff's Alpha from other metrics like correlation for percentage agreement, since it accounts for chance agreement to prevent inflation of agreement values. To understand how $D_e$ is calculated, let us take the following example where we have

\begin{enumerate}
    \item $\mathit{V}$ which is the set of all possible values that coders can assign (e.g., $\mathit{V}=\{1,2,3,4,5\}$ for a 5-point scale).
    \item $n_v$ which is the number of times value $v \in V$ was assigned, across all units and coders.
    \item $N = \sum_{v \in V} n_v$ which is the total number of value assignments.
\end{enumerate}

From this, we can compute the relative frequencies (probabilities):

\begin{equation}
    p_v = \frac{n_v}{N}
\end{equation}

Now we compute the expected disagreement by summing over all unordered pairs of values $(v, v') \in V \times V$, weighted by how frequently those value pairs would co-occur by chance if coder assignments were independent.

We define a weighting term:

\begin{itemize}
    \item If $v \neq v'$, the probability of randomly drawing the pair $(v,v')$ is $2 \cdot p_v \cdot p_{v'}$. The factor of 2 ensures we're counting both $(v,v')$ and $(v',v)$ as one unordered pair.
    \item If $v=v'$, the probability of drawing $(v,v)$ is $p_{v}^2$
\end{itemize}

Using a distance function $\delta (v,v')$, the expected disagreement is:

\begin{equation}
    D_e = \sum_{v \in V} \sum_{v' \in V} p_v p_{v'}\, \delta(v, v')
\end{equation}

Conceptually, this represents the expected value of the disagreement between two randomly selected values from the overall distribution.

When observers agree perfectly, $D_o=0$ and $\alpha=1$, indicating perfect reliability. When observers agree only by chance, $D_o=D_e$ and $\alpha=0$, indicating absence of reliability.

\subsection{Distance Functions}

A crucial aspect of Krippendorff’s alpha is the distance function $\delta(v,v')$, which quantifies how much two coded values disagree. The choice of $\delta$ reflects the measurement level of the data and fundamentally affects the computation of disagreement. In general $\delta$ must satisfy $\delta(v,v') = 0$ when two raters agree and $\delta(v,v') > 0$ otherwise. Different standard choices of $\delta$ are used for nominal, ordinal, and interval data. Each distance function changes how disagreement is accumulated in $D_o$ and $D_e$.

\subsubsection{Nominal Distance} For nominal (categorical) data with no inherent order, the standard distance function is the discrete (binary) metric. In this case, any two different categories are simply considered maximally different. Formally, one sets

\begin{equation}
\delta_{\text{nominal}}(v, v') = 
\begin{cases}
  0, & v = v' \\
  1, & v \neq v'
\end{cases}
\end{equation}

Therefore, exact matches incur zero distance and any mismatch contributes a distance of 1.  Using this $\delta$ means the observed disagreement $D_o$ is effectively the count (or weighted count) of all pairwise coding disagreements among raters.

To calculate $D_o$ using nominal distance,

\begin{equation}
    D_o = \sum_{\substack{\text{pairs}}} \delta_{\text{nominal}}(v_i, v_j) 
\end{equation}

So $D_o$ is essentially the count of mismatches.

Nominal distance is appropriate when values are simply labels or categories (e.g. color names, types of object, or coding categories like “yes/no”, “red/green/blue” etc.). Here there is no notion of “how different” two distinct categories are, only that they are different. In effect, $\alpha$ with nominal $\delta$ reduces to a chance-corrected proportion of exact-agreement (similar in spirit to kappa). All disagreements are weighted equally, so a mild coding error (“cat” vs “dog”) counts the same as a gross one (“cat” vs “car”) in $D_o$. This is the metric we used in the SummaC benchmark, since the items there have binary labels.

\subsubsection{Ordinal Distance} For ordinal data (ranked categories, e.g. survey responses “low/medium/high” or Likert scales), the distance function must respect the ordering of values. A common choice is to use the squared difference in ranks (or a formula based on cumulative frequencies of ranks). Krippendorff’s original formulation defines the ordinal distance as

\begin{equation}
    \delta_{\mathrm{\mathit{ordinal}}}(v, v') = 
\left(
    \sum_{g = \min(v, v')}^{\max(v, v')} n_g
    - \frac{n_v + n_{v'}}{2}
\right)^2
\end{equation}

Here, $n_v$ is the frequency of category $\mathit{v}$.  In simpler terms, this usually reduces to the squared difference in rank position between vv and v'v', possibly standardized by category frequencies. The key effect is that $\delta_{\rm \mathit{ordinal}}(v,v')$ increases as the categories are farther apart in rank.

For ordinal $\delta$, if $N$ units are coded and unit $j$ has $m_j$ coders giving values $v_{1j},\dots,v_{m_j j}$, then the total number of coder-pairs is

\begin{equation}
    n = \sum_{j=1}^{N} \binom{m_j}{2}
\end{equation}

Using the ordinal distance $\delta_{\rm \mathit{ord}}$, Krippendorff’s $D_o$ can be written as

\begin{equation}
    D_o = \frac{1}{n} \sum_{j=1}^{N} \sum_{i < i'} \delta_{\mathrm{\mathit{ord}}}(v_{ij}, v_{i'j})
\end{equation}

Here each inner sum runs over all $\binom{m_j}{2}$ distinct pairs of coders $(i,i')$ within unit $j$. Equivalently (and as shown by Krippendorff), this is the weighted average of within-unit disagreements. In other words, $D_o$ is the mean of all squared ordinal distances between pairs of ratings of the same unit.

Each category has a rank (e.g.\ $1,2,3,\dots$). The standard ordinal distance (squared) between two categories $c$ and $k$ is defined by how far apart their ranks lie in the observed data distribution. Specifically

\begin{equation}
    \delta_{\mathrm{\mathit{ord}}}(c, k) = 
\left(
    \sum_{g = \min(c, k)}^{\max(c, k)} n_g
    - \frac{n_c + n_k}{2}
\right)^2
\end{equation}

where $n_g$ is the frequency of category $g$ in the pooled data. Intuitively, $\delta_{\rm ord}(c,k)$ counts the total number of cases between $c$ and $k$ (plus half-counts of the endpoints), then squares that gap. Plugging this into $D_o$ means each pair of ratings contributes the square of the rank-gap between their categories.

Ordinal distance is appropriate when categories are ordered but not equally spaced (e.g. ratings like "good, better, best" or "strongly agree, agree, neutral, disagree, strongly disagree"). Using this distance in $\alpha$ means that two raters who give adjacent ranks (e.g. 2 vs. 3 on a five-point scale) incur less disagreement than two raters who give opposite ends (e.g. 1 vs. 5). In formula terms, $D_o$ will sum the squared rank differences for each pair. Thus $\alpha$ will penalize large rank disagreements more heavily. In practice, this often yields a larger $D_o$ (hence smaller $\alpha$) than the nominal distance would, because it encodes more information about how coders differ, not just that they differ. (One can also standardize ordinal distances to lie between 0 and 1, but the relative weighting is what matters for $\alpha$.) This is the distance we used in the SummEval and MTBench benchmarks, since the former has labels on a Likert Scale, and the latter has three labels, $\mathit{model_a}$, $\mathit{model_b}$ and $\mathit{tie}$, allowing us to distinguish between disagreements where get ($\mathit{model_a}$, $\mathit{tie}$) versus ($\mathit{model_a}$, $\mathit{model_b}$).

\subsubsection{Interval Distance} For interval data (quantitative values on a scale with equal intervals), the usual choice is the squared-difference distance. That is, one sets

\begin{equation}
    \delta_{\mathrm{\mathit{interval}}}(v, v') = (v - v')^2
\end{equation}

This simply treats the numerical values as points on the real line and measures squared distance between them.

To calculate $D_o$ from interval distance,

\begin{equation}
    D_o = \sum_{\substack{\text{pairs}}} (v_i - v_j)^2
\end{equation}

So $D_o$ is the sum of squared deviations across all coder pairs.

Interval distance is used when the data are measured on an interval scale (e.g. temperature in Celsius, or any numeric rating where differences are meaningful). For example, if raters assign values 3.0 and 4.5 to an item, $\delta = (3.0-4.5)^2 = 2.25$. In $\alpha$’s computation, each such pairwise difference contributes to the total disagreement. The squared form makes $\alpha$ analogous to a variance-based agreement measure: larger numerical discrepancies inflate $D_o$. In effect, with interval $\delta$, Krippendorff’s alpha becomes sensitive to the magnitude of disagreements. Two coders who differ by 1 unit vs. two coders who differ by 5 units will have drastically different contributions to $D_o$. This is appropriate when numerical differences carry substantive meaning. We do not use this metric in this paper, since none of the datasets in our studies contain numerical labels, however, we are still including this distance function in our explanation for the sake of completeness.

\section{More Details on SummaC Benchmark}
\subsection{Dataset Statistics of SummaC Benchmark}
\label{app:summac_stats}
Table~\ref{tab:summac-stats} shows the number of test examples in each of the six datasets in the SummaC benchmark.
\begin{table}[htbp]
\centering
\small
\begin{tabular}{@{}ll@{}}
\toprule
\textbf{Dataset} & \textbf{Test Examples} \\ \midrule
CoGenSumm~\citep{falke-etal-2019-ranking}        & 400                    \\
XSumFaith~\citep{maynez-etal-2020-faithfulness}        & 1250                   \\
Polytope~\citep{huang-etal-2020-achieved}         & 634                    \\
FactCC~\citep{kryscinski-etal-2020-evaluating}           & 503                    \\
SummEval~\citep{summeval}         & 850                    \\
FRANK~\citep{pagnoni-etal-2021-understanding}            & 1575                   \\ \bottomrule
\end{tabular}%
\caption{Number of Test Examples in Each Dataset in the SummaC Benchmark}
\label{tab:summac-stats}
\end{table}

\subsection{Impact of CoT and Few-shot Prompting on Intra-Rater Reliability and Accuracy on SummaC}
\label{app:summac_prompting}
Two breakthrough advances in the use of language models are few-shot prompting~\citep{fewshot} and chain-of-thought (CoT) prompting~\citep{cot}. Few-shot prompting refers to providing the model with a few labeled examples demonstrating how to perform a desired task with higher accuracy. On the other hand, CoT prompting enables language models to generate intermediate reasoning steps that lead to the desired answer. They represent different facets of in-context learning, and can be combined or used independently.

In our next set of experiments, we investigate whether leveraging these techniques addresses self-reliability. In the few-shot setting, we add 5 positive and 5 negative examples in the prompt. In the chain-of-thought setting, we prompt the model to think step by step before answering. Under the Both setting, we employ both few-shot and chain-of-thought prompting.

\begin{table}[]
\small
\centering
\begin{tabular}{@{}llll@{}}
\toprule
Setting     & Llama 3.1 & Deepseek-R1 & Qwen 3 \\ \midrule
Default     & 0.3263    & 0.6278      & 0.7883 \\
w/ Few-shot & 0.3245    & 0.6166      & 0.7804 \\
w/ CoT      & 0.3219    & 0.6132      & 0.7796 \\
w/ Both      & 0.3206    & 0.6134      & 0.7801 \\\bottomrule
\end{tabular}
\caption{Self-Reliability of LLM judges on SummaC for the default setting, as well as when few-shot and chain-of-thought prompting are used together and separately.}
\label{tab:selfcon_improve}
\end{table}

\begin{table}[]
\small
\centering
\begin{tabular}{@{}llll@{}}
\toprule
Setting     & Llama 3.1 & Deepseek-R1 & Qwen 3 \\ \midrule
Default     & 61.4      & 72.3        & 80.6   \\
w/ Few-shot & 62.9      & 72.8        & 80.6   \\
w/ CoT      & 60.8      & 71.6        & 80.4   \\
w/ Both     & 60.1      & 72.1        & 80.3   \\ \bottomrule
\end{tabular}
\caption{Balanced Accuracy (majority-vote) between the LLM and human judgments for the default setting, as well as when few-shot and chain-of-thought prompting are used together and separately.}
\label{tab:acc_improve}
\end{table}

Table~\ref{tab:selfcon_improve} shows the self-reliability of LLM judges measured by Krippendorff's Alpha over 3 runs. We see that few-shot prompting does not appreciably change the self-reliability of the models, and leads to a slight decrease in consistency. This change is not significant and shows that one cannot easily address the self-reliability issue with different prompting strategies.

Table~\ref{tab:acc_improve} shows the accuracy of LLM judges against human judgments. While few-shot prompting does lead to a small improvement in Llama 3.1 and Deepseek-R1, there is no significant jump in performance for either prompting strategy. This could be because, as previously observed, few-shot prompting does not always work well with reasoning models like Deepseek-R1, Qwen 3 and o1~\citep{deepseekai2025deepseekr1incentivizingreasoningcapability,nori2024from}, while chain-of-thought prompting does not necessarily improve the accuracy of a reasoning model since it already performs chain-of-thought implicitly~\citep{cot_reasoning}. Interestingly, Qwen 3 shows the least change under these new settings, suggesting newer reasoning models are fairly robust to different prompts.

\section{Additional Analysis of Agreement Between Judges in SummEval}
\label{app:summeval}

\begin{figure*}[htbp]
  \centering
  \begin{minipage}{0.17\textwidth}
    \centering
    \includegraphics[width=\linewidth]{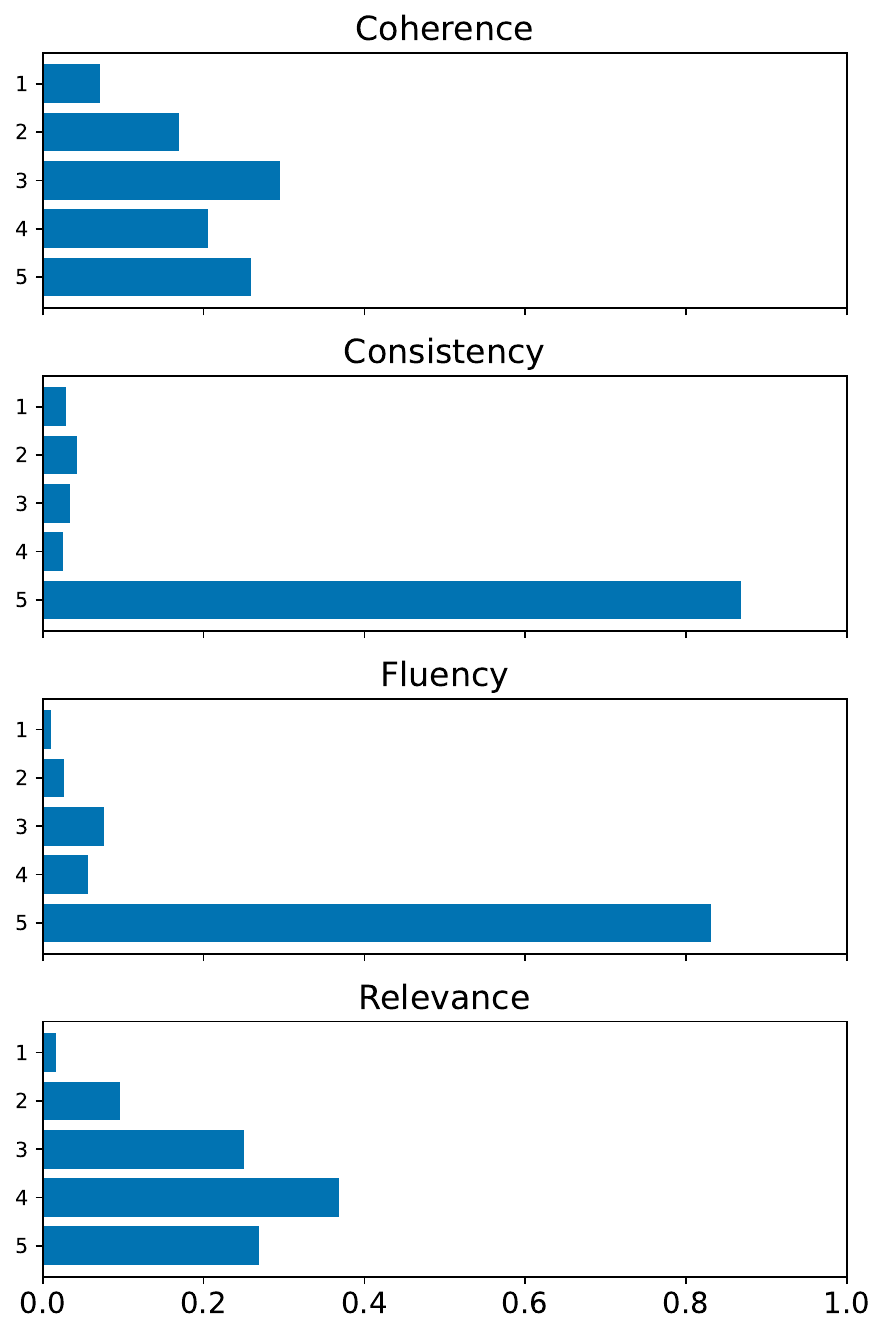}
    \caption*{(a) Experts}
  \end{minipage}
  % \hfill
  \begin{minipage}{0.17\textwidth}
    \centering
    \includegraphics[width=\linewidth]{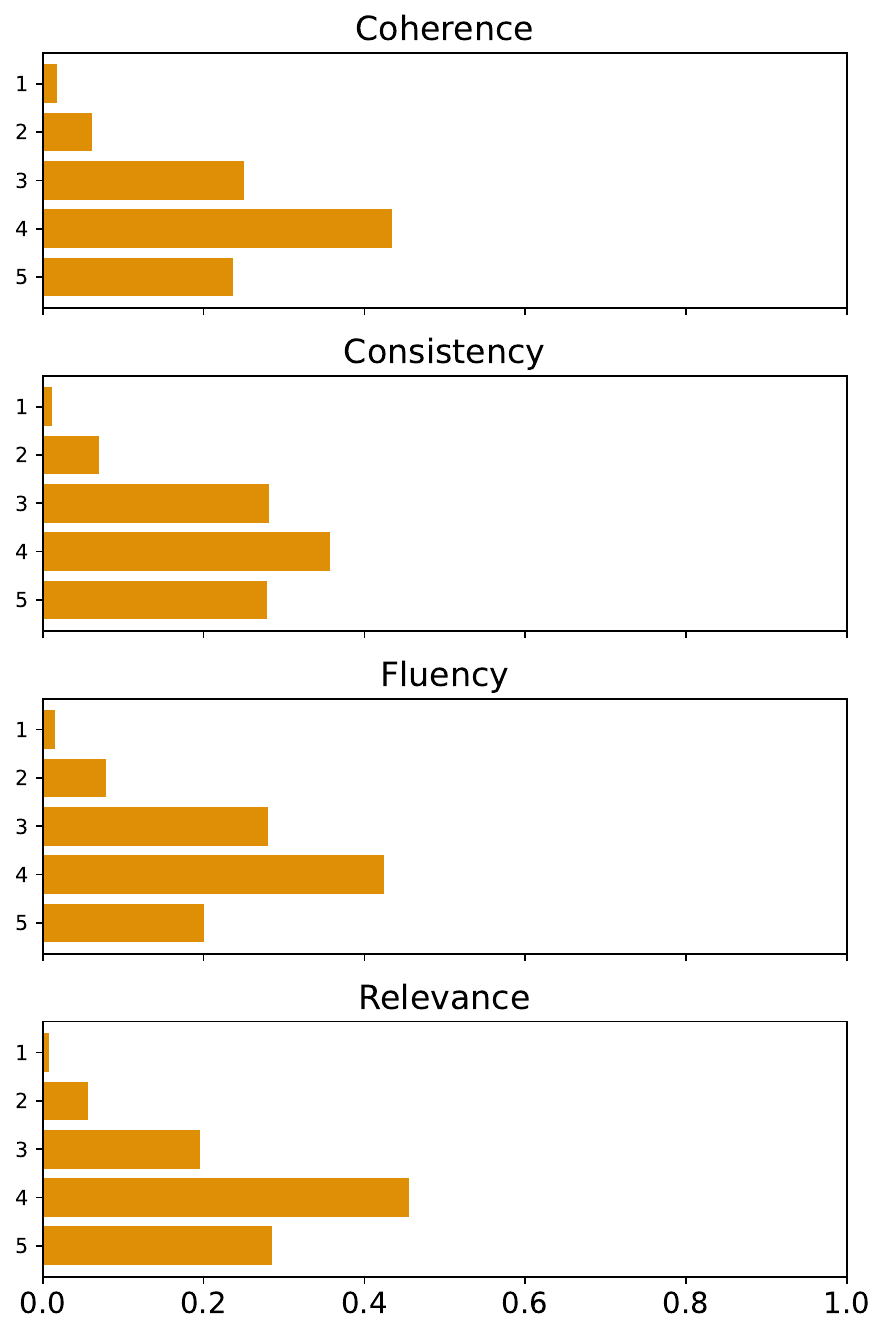}
    \caption*{(b) Turkers}
  \end{minipage}
  % \hfill
  \begin{minipage}{0.17\textwidth}
    \centering
    \includegraphics[width=\linewidth]{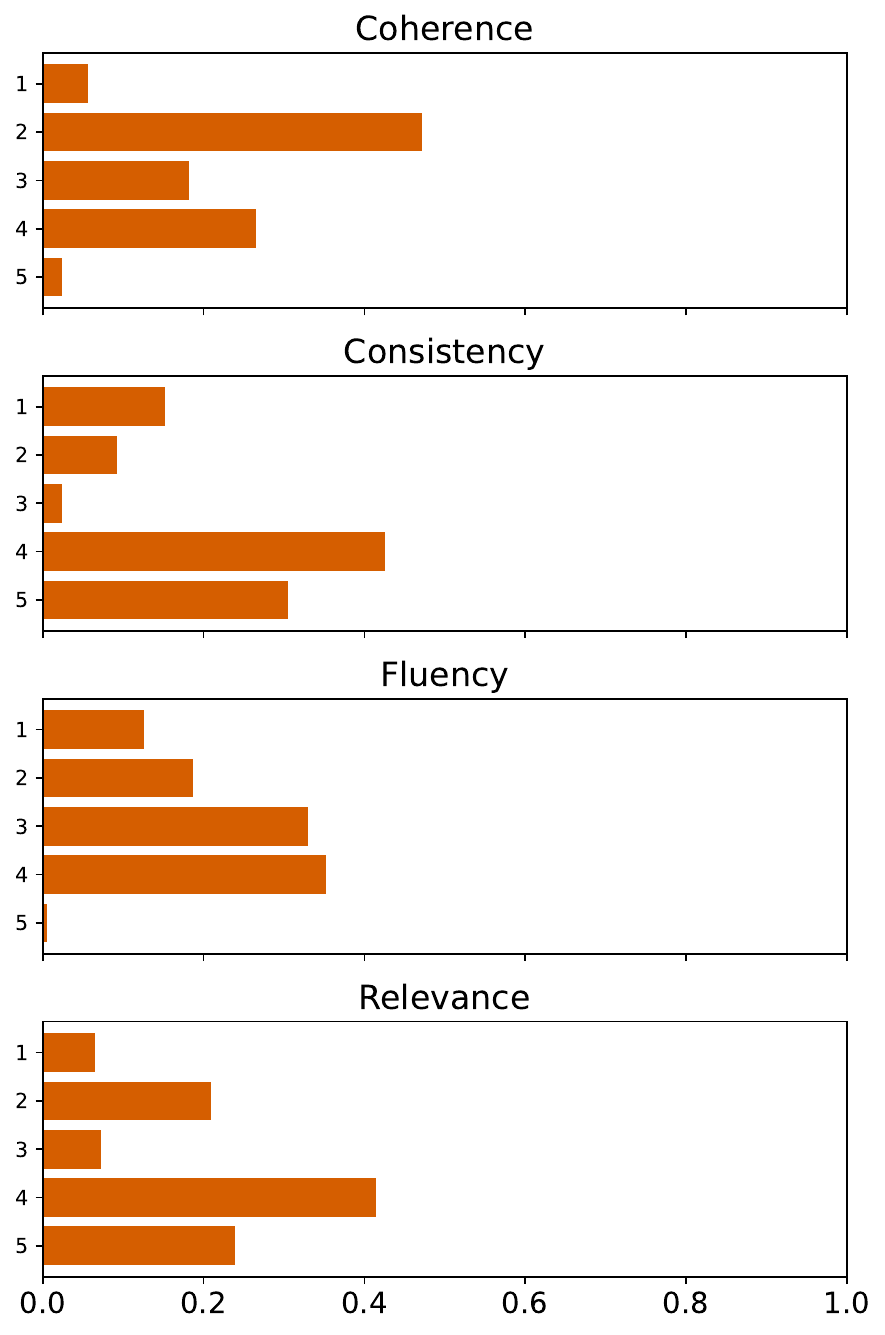}
    \caption*{(c) Llama 3.1}
  \end{minipage}
  % \hfill
  \begin{minipage}{0.17\textwidth}
    \centering
    \includegraphics[width=\linewidth]{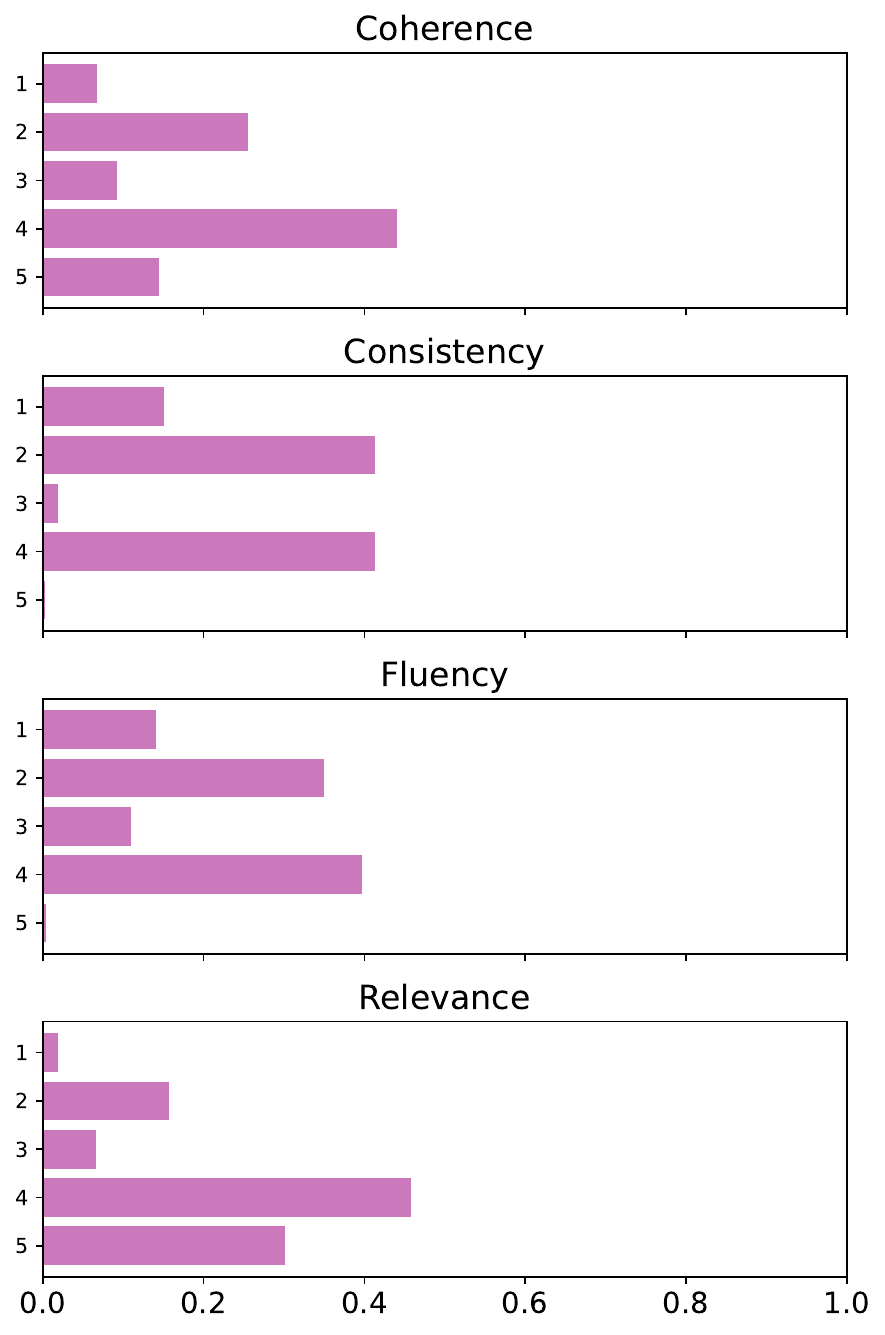}
    \caption*{(d) Deepseek R1}
  \end{minipage}
  % \hfill
  \begin{minipage}{0.17\textwidth}
    \centering
    \includegraphics[width=\linewidth]{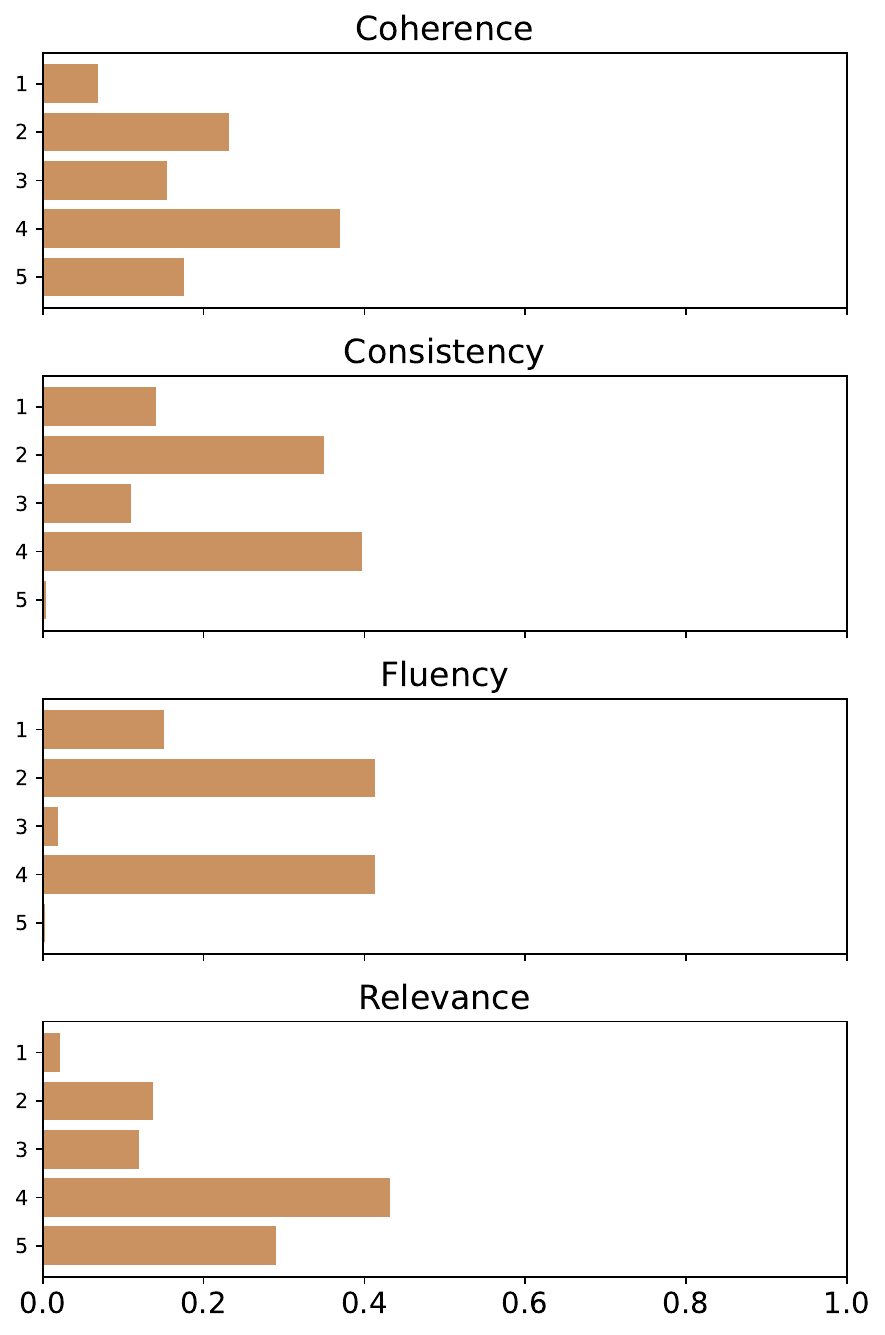}
    \caption*{(d) Qwen 3}
  \end{minipage}

  \caption{Distributions of scores assigned by different raters across all metrics in the SummEval dataset.}
  \label{fig:summeval_score_dist}
\end{figure*}

\subsection{Distributions of Scores in SummEval From Different Populations of Raters}
\label{app:score_dist}
In Figure~\ref{fig:summeval_score_dist}, we see the distributions of scores assigned by different categories of raters across all four metrics on the SummEval benchmark. We see that for Turkers the distributions do not change appreciably across metrics, indicating that human judges without proper expertise do a poor job of differentiating between different metrics. While the other four categories of judges do discern between metrics, they all have very different distributions of scores for a specific metric. Even though Deepseek R1 and Qwen 3 showed much higher agreement with experts compared to LLama 3.1 and Turkers, we see that their distributions are very different. This implies that each category of judge follows their own internal definition of a given metric, which puts a strict cap on how high their agreement can be.

\section{Additional Details on MT-Bench Benchmark}
\subsection{Breakdown of Examples in MT-Bench}
\label{app:mtbench_examples}
\begin{table}[htbp]
\centering
\begin{tabular}{rrrr}
\hline
2 & 3 & 4 & 5\\
\hline
599 & 132 & 24 & 6\\
\hline
\end{tabular}
\caption{Number of examples with 2, 3, 4, or 5 judgments from human annotators in MT-bench.}
\label{tab:mtbench_count_judges}
\end{table}

Table~\ref{tab:mtbench_count_judges} shows how many examples had 2, 3, 4, or 5 human judgments in the MT-Bench dataset. These are the examples we used in our experiments since we filtered out examples that had only one human judgment.

\end{document}